# A Glimpse in ChatGPT Capabilities and its impact for AI research

Frank Joublin[1], Antonello Ceravola[1], Joerg Deigmoeller[1], Michael Gienger[1], Mathias Franzius[1], Julian Eggert[1]

[1] Honda Research Institute Europe

**Abstract**. Large language models (LLMs) have recently become a popular topic in the field of Artificial Intelligence (AI) research, with companies such as Google, Amazon, Facebook, Amazon, Tesla, and Apple (GAFA) investing heavily in their development. These models are trained on massive amounts of data and can be used for a wide range of tasks, including language translation, text generation, and question answering. However, the computational resources required to train and run these models are substantial, and the cost of hardware and electricity can be prohibitive for research labs that do not have the funding and resources of the GAFA. In this paper, we will examine the impact of LLMs on AI research. The pace at which such models are generated as well as the range of domains covered is an indication of the trend which not only the public but also the scientific community is currently experiencing. We give some examples on how to use such models in research by focusing on GPT3.5/ChatGPT3.4 and ChatGPT4 at the current state and show that such a range of capabilities in a single system is a strong sign of approaching general intelligence. Innovations integrating such models will also expand along the maturation of such AI systems and exhibit unforeseeable applications that will have important impacts on several aspects of our societies.

## 1. Scope and Purpose

Large Language Models (LLMs) are complex artificial intelligence models that use natural language processing (NLP) techniques to analyze, understand, and generate human-like language. LLMs can learn from vast amounts of data, enabling them to generate human-like responses to a variety of prompts, including text, images, and audio. Although several LLMs are available today, such as PaLM [Chowdhery et Al. 2022] or LaMDA [Thoppilan et Al. 2022] from Google, Sparrow [Glaese et Al. 2022] from Deepmind, BlenderBot 3 [Shuster et Al. 2022] or Galactica [Taylor et Al. 2022] from Meta, or Megatron [Shoeybi et Al. 2019] from NVIDIA to cite a few, this paper will focus on ChatGPT [OpenAI 2022] from OpenAI, a conversational AI that has raised a lot of scientific and media interest in recent months. ChatGPT is designed to generate text-based responses to natural language prompts, and it has demonstrated remarkable success in several areas.

The purpose of this paper is to provide a comprehensive overview of LLMs and the capabilities of ChatGPT in particular. The paper aims to give examples of creative prompt design to drive LLMs toward specific applications, such as natural language understanding, reasoning, knowledge manipulation, common sense, as well as planning, exploration, or interactive games. Additionally, the paper seeks to discuss the impact of LLMs on AI research.

One of the primary objectives of this paper is to highlight the versatility of LLMs and their ability to be trained in a wide range of tasks. LLMs have been utilized to create poetry, answer trivia questions, and even write entire novels, and their potential as a general AI has never been closer. Specifically, the extent to which their capabilities can be pushed to integrate with perception and memory in larger systems is yet to be determined.

## 2. Introduction

LLMs are one class of deep learning models that are primarily designed to predict, complete and generate snippets of natural language text, but they are increasingly also used for text-based general comprehension tasks. Recent research has shown that LLMs are highly effective in a wide range of natural language tasks [Qin et Al. 2023], such as generating condensed versions of texts, machine translation, text classification [Minaee et Al. 2021], sentiment analysis [Susnjak 2023], question-answering, misinformation detection, task-oriented dialogue, storytelling [Nichols et Al. 2020], knowledge grounded dialogues [Bang et Al. 2023], named entity recognition [Qin et Al. 2022], and code generation [Sadik et Al. 2023]. Most of these impressive performances are considered to be emergent properties appearing with size scaling [Kaplan et Al. 2020]. Another type of deep learning models, which have raised large media attention, are Large Diffusion Models (LDM) which allow the creation of high-resolution images from textual prompts [Yang et Al. 2022] [Gozalo-Brizuela et Al. 2023]. LDMs are based on a similar architecture to LLMs for their textual part. LLMs excel in these tasks due to two crucial factors. Firstly, LLMs utilize transformers, a cutting-edge neural network architecture, a self-attention mechanism, and a huge set of parameters, which enables the model to understand the relations between various input elements. Secondly, LLMs employ a two-stage training process [Radford et Al. 2018] to learn from data in an efficient manner. During the initial pre-training stage, LLMs use self-supervised learning to learn from vast amounts of data without requiring any form of annotation. This feature is a significant advantage over traditional fully supervised deep learning models, as it eliminates the need for extensive manual annotation, making it more scalable. In the subsequent fine-tuning stage, LLMs are trained on smaller, task-specific, annotated datasets to leverage their knowledge obtained during the pre-training stage to perform specific tasks as intended by end users [Glaese et Al. 2022] [Ouyang 2022]. As a result, LLMs achieve high accuracy on various tasks with minimal human intervention.

### 2.1. Short description of ChatGPT Architecture

Figure 1 depicts the architecture of ChatGPT3.5 [OpenAI 2022] in a schematic view. The model is based on a decoder-only structure that has been proven to be sufficient for language generation. From the bottom left of the figure, the system uses a technique called tokenization, which involves breaking down the input text into smaller units known as tokens. A token represents a part of a word that is approximately four characters long. These tokens are then mapped to vectors in a 1024-dimensional space using a dictionary of around 40,000 token entries, allowing the model to process them efficiently. The vector embedding size is a crucial parameter in the transformer architecture because it determines the dimensionality of the embedding space in which the tokens are represented. A higher embedding size can help the model capture more fine-grained information about the tokens, but at the cost of computational complexity. The vector embedding step also defines the semantic space that will be propagated and transformed along the different layers of the architecture.

The architecture consists of 96 transformer blocks stacked on top of each other, forming a huge feed-forward network. Each block contains a multi-head attention mechanism and a feed-forward neural network, as shown in the right part of Figure 1. The multi-head attention mechanism [Vaswani 2017] allows the model to attend to different parts of the input sequence simultaneously, while the feed-forward neural network applies a non-linear transformation to the input data and brings the output back to the semantic space dimensionality for each token position. This allows each decoder to have the same input and output dimensionality and enables scaling by stacking decoders on top of each other.

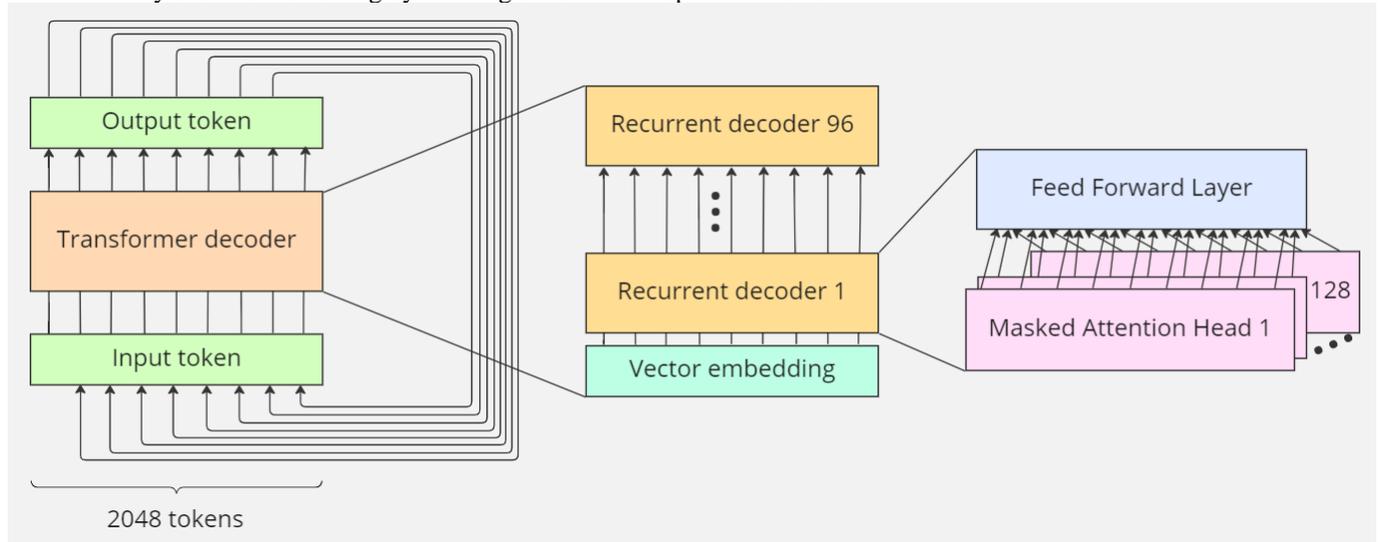

Fig. 1: Schematic view of ChatGPT architecture

In the last layer of the architecture (top-left of Figure 1), the model uses a SoftMax function to select the most likely token for the next position in the sequence. The SoftMax function takes the output of the model at the current position and produces a probability distribution over the entire vocabulary. Each token in the vocabulary dictionary is assigned a probability, and a token with probability among the highest is selected as the next token in the sequence. After a token is selected, it is fed back into the model in an auto-regressive way. This means that the selected token is appended to the input sequence, and the model generates the next token in the sequence based on the updated input. The process is repeated for each token in the sequence, with the model generating one token at a time.

The basic architecture of GTP-4 has not been disclosed by OpenAI. However, it includes multimodal inputs, such as vision, and contains up to 1000 times more parameters than its predecessor. Additionally, the context window has been increased by 16 times, allowing for larger output production and improved accuracy. To make it safer and more robust, a post-treatment was applied, which took hundreds of associates and external specialists during six months of work, as described in [OpenAI 2023]. For further information on the differences between the two versions, please refer to [Koubaa 2023].

**2.2. Do LLMs understand texts?**

Each attention layer in ChatGPT is composed of three linear transformations: the query, key, and value transformations. The query transformation takes the current semantic vector at a given position as input and transforms it into a set of query vectors. The key transformation transforms each semantic vector from the context into a set of key vectors. These semantic vectors are then used to compute attention weights based on proximity (dot product) between queries and keys. The resulting attention weights are used to create the value vectors, which are the output of the attention layer, by a weighted sum of the context. During training, these transformations are specialized to specific semantic transformations gated by the current token. Such transformation of the semantic space enables the model to reason and generate meaningful text by building on the contextual information provided by previous words in a sentence. Successive transformations of the semantic space allow ChatGPT to make more complex and nuanced predictions based on the accumulated contextual information. In fact, transformer architectures using attention mechanisms have been shown to be Turing complete [Pérez et al., 2021].

For example, consider the following sentence: "I went to the store to buy some apples, but they were all out." When ChatGPT processes this sentence, it begins by transforming the semantic space to represent the meaning of each word in the sentence. As it progresses through the sentence, it updates the semantic space based on the context of each word. At the point where the word "apples" appears, the model activates a semantic vector encoding that "apples" are a type of fruit that can be bought at a store. When it encounters the word "but" in the next part of the sentence, it updates the semantic space to reflect the contrast between the expectation of being able to buy apples at the store and the reality that the store was out of apples. The resulting semantic vectors generated at each layer represent the whole meaning of the sentence step by step.

This ability to reason and understand the meaning of words in context is what makes ChatGPT a powerful tool for natural language processing tasks such as text generation, translation, and summarization. By building on the accumulated context provided by previous words in a sentence, the model can make more sophisticated predictions and generate text that is more coherent and meaningful. Since these semantic vectors represent meaning in a high-dimensional space (1024 dimensions), it is highly probable that these vectors encode "understanding" in a form that is highly opaque to human description. Nevertheless, it would be hazardous to consider that LLMs do not understand the text or task they manipulate. A recent paper [Li et al., 2022] explores the abilities of language models and their reliance on internal representations by using a GPT model to predict legal moves in Othello. They find evidence of an emergent nonlinear internal representation of the board state and use intervention techniques to produce "latent saliency maps" that help explain predictions.

### 2.3. Comparison with traditional form of reasoning

Traditional reasoning systems rely on formal representations of knowledge and a set of rules or procedures for manipulating that knowledge to arrive at a solution. These systems are often used in areas such as game playing, planning, and decision making, where a set of well-defined rules and objectives can guide the search for a solution in a state space.

The key advantage of language models like ChatGPT is their ability to learn and generalize from large amounts of data. They are not restricted to a specific set of rules or procedures but can adapt to different contexts and tasks based on the patterns they learn from the input data. This makes them highly flexible and adaptable, allowing them to generate responses that are more natural and human-like than traditional rule-based systems. Language models reason in a feedforward way, based on patterns and semantic transformations learned from data. This speeds up their responses since no search in a huge space is needed, but it also makes them struggle with tasks that require more specialized knowledge, such as embodied knowledge or goal-directed reasoning abilities [Binz & Schulz, 2023].

## 3. Use-Cases I (in isolation)

A straightforward way to use LLM's is to use them as a data providing module or a "translation" module within a feedforward information flow scheme. For example, ChatGPT receives some input text, and generates a response as output.

In the following subsections we describe potential and tested use cases where ChatGPT was used in this way.

### 3.1. Natural Language Usage

#### 3.1.1. Text Generation with Tone

As a preprocessing step before a text-to-speech interface, LLM can be used to reformulate sentences in a specific desired tone, such as "joking", "depressed", or "formal". Here is an example of a prompt that could be used in a robotic scenario, where the same phrase has to be produced in various moods depending on the context (Fig. 2)

```
PROMPT: I will give you sentences preceded by a mood or tone indication in the format [mood indication] <sentence>.
Reformulate the sentence using the mood indication. just output the reformulation without extra comments. Are you ready?
AI: Yes, I'm ready. Let's get started.

PROMPT: [joking] I found the ball in the drawer.
AI: Just kidding, I found the ball in the drawer.

PROMPT: [depressed] I found the ball in the drawer.
AI: Sadly, I found the ball in the drawer.

PROMPT: [formal] I found the ball in the drawer.
AI: I have discovered the ball in the drawer.

PROMPT: [polite] I found the ball in the drawer.
AI: Excuse me, I found the ball in the drawer.

PROMPT: [sorry] I found the ball in the drawer.
AI: I'm sorry, but I found the ball in the drawer.

PROMPT: [laughing] I found the ball in the drawer.
AI: Haha, I found the ball in the drawer.

PROMPT: [surprised] I found the ball in the drawer.
AI: I can't believe it, I found the ball in the drawer!

PROMPT: [very formal to a professor] I found the ball in the drawer.
AI: Excuse me, Professor, I have found the ball in the drawer.

PROMPT: [hiding the truth] I found the ball in the drawer.
AI: Well, I'm not sure how it happened, but the ball ended up in the drawer.
```

Fig. 2: Example of ChatGPT3.5 output in a reformulation task

### 3.1.2. Natural Language Preprocessing

Since LLMs are designed for language processing, we would expect them to excel at extracting "meaningful" components from text. This is where Natural Language Preprocessing (NLP) comes in, which includes tasks such as slot filling, intent classification, and reference resolution. The objective is to identify specific parts of a text, referred to as "slots," that belong to a particular "type" (e.g., the object or the agent of a sentence) and are assumed to contain essential meaning. This is one of the first steps for any Natural Language Understanding (NLU) system (see Fig. 3). For further reading see [Eggert 2022].

```
PROMPT: Extract the following semantic components from the following sentence:
- The main agents executing an action
- The type of actions being executed
- The main objects or subjects on which the action is executed
- The main recipients of the action result
Return a json-structure with the results

Sentence: "Tim serves a cold drink to Eve."
AI:
{
    "Agents": ["Tim"],
    "Action": ["serves"],
    "Object/Subject": ["cold drink"],
    "Recipient": ["Eve"]
}
PROMPT: Sentence: "Tim and Tom bring and serve a cold drink and snacks to Eve and Ana."
AI:
{
    "Agents": ["Tim", "Tom"],
    "Action": ["bring", "serve"],
    "Object/Subject": ["cold drink", "snacks"],
    "Recipient": ["Eve", "Ana"]
}
```

Fig. 3: Example of ChatGPT3.5 output in a NLU task

One subtask, known as slot filling, involves identifying and returning the parts of a sentence that constitute slots of a specific type. Large Language Models (LLMs) serve as formal translators in this task, converting plain natural language into a formatted, annotated text (refer to Fig. 4).

```
PROMPT: In the following sentences identify the agent, the action and the object:
    Phrase: Could you give me the pencil?
AI: Agent: You
    Action: give
    Object: the pencil
PROMPT: Phrase: Check the book!
AI: Agent: You (implied)
    Action: Check
    Object: the book
PROMPT: Phrase: Pretty Sue drives the red car along the road.
AI: Agent: Pretty Sue
    Action: drives
    Object: the red car
PROMPT: Phrase: The red car is driven along the road by pretty Sue.
AI: Agent: Pretty Sue
    Action: drives (passive voice)
    Object: the red car
PROMPT: Phrase: The chicken is ready to eat.
AI: Agent: None (implied: someone who cooked the chicken)
    Action: is ready
    Object: the chicken
PROMPT: Is there another interpretation of the last sentence?
AI: "The chicken is ready to eat" could also be interpreted as the chicken being in a state of readiness to eat something else, rather than being ready to be eaten itself. However, this would be an unusual interpretation in most contexts, as the more common and straightforward meaning is that the chicken has been cooked and is now ready for someone to eat it.
PROMPT: for this other interpretation of the sentence identify the agent, the action and the object:
AI: Sure, here's how we could interpret the sentence under the alternative interpretation:
    Agent: The chicken (or the person feeding the chicken)
    Action: is ready
    Object: to eat (something else)
```

Fig. 4: Example of ChatGPT3.5 output in a NLU task

This is very similar to the formal decomposition of text into its syntactic components, as done in school when identifying the verb, subject, and object of a given sentence. The traditional method of doing this is e.g. by use of syntactic parse trees. However, these methods usually underperform when language ambiguities are present. Similarly, ChatGPT occasionally exhibits errors in such cases. Interestingly, we can readily convince ChatGPT to generalize and return the agent of an action, rather than the linguistic "subject" of a verb. This enables generalization over active and passive constructions. For example, in "Tom drives the car" and "The car is driven by Tom," the syntactic subject is first Tom and then the car, whereas ChatGPT returns Tom in both cases as the semantically relevant agent performing the action.

Another subtask is to resolve indirect references to meaningful text components, which are expressed in language by proxy words such as "he", "it", and "there", to their underlying meanings (i.e. who is being referred to by "he", what is meant by "it", and where exactly is "there"?). In the following example, we can see that ChatGPT is very good at performing this task (see Fig.

5). This is important for comprehending larger passages of text and usually requires dedicated machine learning or designed machinery. Thus, ChatGPT could be used as a very compact and powerful replacement for such machinery.

```
PROMPT: Take a text and tag all first appearances of people, agents, objects, places and times with a number in python dictionary format.  Then replace
all indirect references to people ("he", "she", "they", "his", "her", "you"), agents ("it", "its", "them", "their"), objects ("it", "them"), places
("here", "there") and times by their number tag.

Input: Tim drives the car to the washing store. There, he gets out and also refuels. It is really dirty.
AI: Tagging: {1: "Tim", 2: "the car", 3: "washing store"}
    Output: 1 drives 2 to the 3. At the 3, 1 gets out and also refuels. 2 is really dirty.

PROMPT: Input: Tim and Tom went to the airport. When they arrived there, their relatives were waiting for them.
AI: Tagging: {1: "Tim", 2: "Tom", 3: "airport"}
    Output. Tim and Tom went to the airport. When 1 and 2 arrived at the airport, 1 and 2's relatives were waiting for 1 and 2.

PROMPT: Input: Tom gave Eve a present during her birthday. She liked it very much then.
AI: Tagging: {1: "Tom", 2: "Eve", 3: "a present", 4: "during her birthday"}
    Output: 1 gave 2 3 4. 2 liked it very much 4.

PROMPT: Input: She drove the car and returned it at midnight. Then she got drunk.
AI: Tagging: {1: "She", 2: "the car", 3: "at midnight"}
Output: 1 drove 2 and returned it 3. 3 1 got drunk.
```

Fig. 5: Example of ChatGPT3.5 output in an indirect reference task

### 3.2. Common-sense Reasoning

#### 3.2.1. Semantic Disambiguation

As ChatGPT was largely trained using semantic, human-curated data sources from the internet, such as WordNet, ConceptNet, and Wikipedia, it is expected that the model is able to reliably answer questions related to encyclopedic knowledge. This is especially relevant for so-called "common-sense knowledge" about items, actions, and properties, which usually needs to be gathered from a variety of sources.

For applications that require to access these data sources, it could be a viable solution to not access each data source separately with its own API, but to "ask" ChatGPT directly. In this sense, LLMs would be used as data extraction resources. One example is the extraction of typical characteristic properties of items, such as "What makes an elephant special?" or "What is the difference between dogs and cats?" In the experiment shown in Figure 6, we demonstrate that ChatGPT can be used in a straightforward way to retrieve this kind of data.

```
PROMPT: Explain the meanings of "chair" in the sentence "The committee chair sat in the center chair.".
AI: In this sentence, "chair" has two meanings. The first meaning is that of a leader or head of a committee. This person is referred to as the
"committee chair" and is the person in charge of organizing and leading the committee. The second meaning of "chair" is a physical piece of furniture.
In this sentence, the committee chair is sitting in the "center chair," which is a chair in the center of the room or table.

PROMPT: From this result, convert the meanings of "chair" to their corresponding WordNet Synset ID of the form word.x.nn. Return as a list of pairs of
Synset ID and explanation.
AI: "Chair" (noun):
1. Chair.n.01: A seat for one person, with a support for the back (synset explanation: a seat for one person, with a back and usually four legs)
2. Chairperson.n.01: The officer who presides at the meetings of an organization (synset explanation: the presiding officer of an organized body of
people)
```

Fig. 6: Example of ChatGPT3.5 output in a direct semantic task

To ensure semantic disambiguation, we investigated whether ChatGPT could comprehend the correct meaning of words in the context of a sentence. In addition to this, we requested ChatGPT to return Wordnet Synsets, as it had some familiarity with them. However, the Synsets returned were mostly incorrect. ChatGPT4 even stated that "As an AI language model, I cannot directly access WordNet". Although this is not a real issue, as OpenAI has recently introduced the possibility of developing "plugins" [OpenAI plugins 2023] that would allow ChatGPT to query specific databases. Unfortunately, at the time of writing, the plugin feature was only available to a limited list of users, and we were unable to test this feature. However, examples such as [Wolfram 2023] demonstrate that this approach works effectively for much more complex problems, and we have no doubt that it could work for retrieving the exact synset from Wordnet.

In our next experiment, we aimed to identify the "typical" (i.e., properties commonly associated with) and "characteristic" (i.e., properties that distinguish it from other concepts) features of items such as objects and agents. Figure 7 displays the results obtained from the task of discriminating between dogs and cats. The findings highlight a significant characteristic of LLMs, which is their ability to retrieve general knowledge and present it in a structured and concise form.

#### 3.2.2. Machine Common-Sense Extraction

Intelligent mobile agents, especially robots that operate in physical environments, require a dedicated type of common-sense knowledge. This knowledge should support an agent in understanding the context of a situation and inferring the necessary steps to fulfill a certain task [Deigmoeller 2022]. In the past, this has been discussed as "machine common sense" [Gunning 2018], which focuses on actions with corresponding agents (who can execute the action), objects (on which the agent applies the action), tools (which can be used to support the action), locations (where the action is performed), and times (when it is performed). Each participant in an action has a defined semantic role.

The semantic roles that determine the context of a physical action are well defined, structured, and studied in the areas of Linguistics [Brown 2019] and Robotics [Zech 2018]. However, a major drawback is the missing knowledge about the common relationships between participants in such action context structures. Large Language Models provide an extremely large amount of such knowledge; unfortunately, this knowledge comes unstructured. Combining both worlds, logical (symbolic) and empirical (sub-symbolic), is a very promising direction.

```
PROMPT: What are specific properties that set apart dogs from mammals? Make a separate list for each of the two.
AI: Dogs:
        1. Domesticated animals
        2. Have a long history of companionship with humans
        3. Can be trained to perform specific tasks
        4. Have a wide variety of breeds
        5. Have a strong sense of smell and hearing
        6. Have a strong pack mentality
        7. Have a high level of loyalty and devotion
    Mammals:
        1. Have a four-chambered heart
        2. Possess sweat glands
        3. Have hair or fur covering their bodies
        4. Have a diaphragm for breathing
        5. Give birth to live young
        6. Have specialized teeth that aid in digestion
        7. Regulate their body temperature internally

PROMPT: Are dogs a social animals? Compare with cats.
AI: Yes, dogs are social animals. They have an instinctive need to be around other animals, both canine and human. In comparison, cats are also social animals, but they are less reliant on their social bonds and tend to be more independent. Cats can form strong bonds with other cats, but they are not as dependent on their social bonds as dogs are.

PROMPT: What are specific physical or appearance properties that set apart dogs from mammals? Make a separate list for each of the two.
AI: Dogs:
        1. Fur or hair coat with a variety of colors and patterns
        2. Floppy ears
        3. Long muzzle
        4. Short, stubby legs
        5. Tail with a curl or plume at the end
        6. Webbed feet
    Mammals:

        1. Hair or fur covering their bodies
        2. Endotherm - regulating body temperature internally
        3. Vertebrae - backbone
        4. Mammary glands for nursing young
        5. Ears and nose for detecting sound and smell
        6. Bodies adapted for running, flying, or swimming
```

Fig. 7: Example of ChatGPT3.5 output in a semantic differentiation task

This section focuses on extracting action-centered knowledge from ChatGPT and representing it in a structured semantic form. The examples presented here are preliminary studies for a more extensive evaluation and comparison of ChatGPT with other language models.

To begin our approach to knowledge extraction, we draw a connection to our internally developed knowledge representation system, MemNet [Eggert 2020]. Utilizing ChatGPT for knowledge extraction, we leverage the concept of action patterns, which are derived from the semantic roles we described earlier. An action pattern consists of a central concept, or action, which links all associated semantic roles (as depicted in Fig. 8). While we have previously extracted subsets of action patterns in prior publications [Losing 2021, Losing 2022], the availability of large language models makes it increasingly feasible to address this challenge at scale.

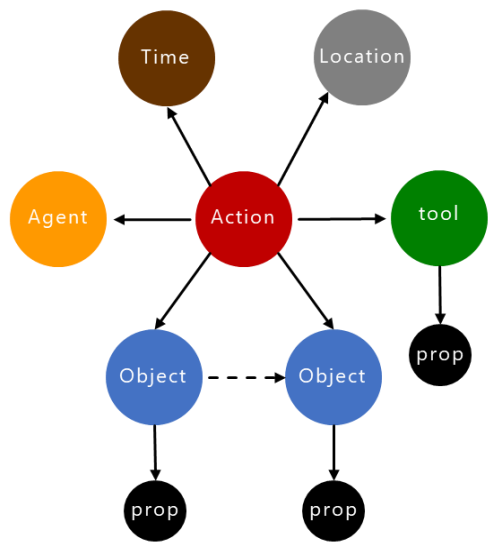

Fig. 8: Action pattern with related semantic roles

In the first experiment, we assume that we already know an action-object relation (represented by the red and blue nodes in Figure 8) and that we want to identify the missing tool relation (represented by the green node in Figure 8). We ask ChatGPT to infer possible tool candidates for the action of "chopping" and the object of "vegetables". Furthermore, we instruct ChatGPT to choose candidates from a provided list of tools and to order them by relevance (as shown in Figure 9).

```
PROMPT: Given the list of object:
['apple', 'banana', 'lime', 'peach', 'plum', 'cereal chips', 'cracker', 'whipped cream', 'sofa', 'kitchen counter', 'coffee table', 'chair', 'dining
table', 'bookshelf', 'flower pot', 'bowl', 'plate', 'frying pan', 'cooking pot, bottle', 'water glass', 'wine glass', 'mug', 'wine', 'milk', 'juice',
'bun', 'bread', 'bread slice', 'fish', 'chocolate syrup', 'tv', 'toaster', 'microwave', 'remote control', 'fridge', 'knife', 'fork', 'pillow', 'key',
'candle', 'faucet', 'vase', 'book', 'orchid', 'tv stand']
Consider each of the objects in this list and return for each object how suited it is in percent for the following usage. Consider also non-standard
way of using each object:
Usage: as a tool to chop vegetables.
Provide relevance as  a percental value. 0% being not relevant, 100% being the best relevance for the question asked.
Provide an answer only in the form of an array of JSON object like: [ { object, relevance, explanation }, ... ] containing the 5 most relevant objects.

AI: Here is an array of JSON objects with the five most relevant objects for using as a tool to chop vegetables and their corresponding relevance
percentages:
[
  {"object": "knife", "relevance": 100, "explanation": "A knife is specifically designed for chopping vegetables, so it is the most relevant object."},
  {"object": "frying pan", "relevance": 10, "explanation": "While a frying pan is not designed for chopping vegetables, it could be used as a makeshift
chopping surface in a pinch. However, it is not very suited for this purpose, so its relevance is low."},
  {"object": "cooking pot", "relevance": 5, "explanation": "Similar to a frying pan, a cooking pot could be used as a makeshift chopping surface, but
it is not designed for this purpose and is not very well-suited."},
  {"object": "bottle", "relevance": 0, "explanation": "A bottle is not at all suited for chopping vegetables."},
  {"object": "tv", "relevance": 0, "explanation": "A TV is not at all suited for chopping vegetables."}
]
```

Fig. 9: Example of ChatGPT3.5 output in a task of tool retrieval

In the next step, we extend our action pattern by considering the consequences of chopping the object, taking into account a list of possible properties. This concept is also known as pre-condition and effect in robotics, which refers to the change in state before and after applying the action of "chopping" (Fig. 10).

```
PROMPT: Given the list of states: ['warm', 'hot', 'cold', 'red', 'green', 'blue', 'black', 'white', 'soft', 'solid', 'fluid', 'full', 'empty', 'on',
'off', 'cut', 'not cut', 'sliced', 'dirty', 'clean']
Consider each of the state in this list and return for each state how probable it can be in percent to characterize the state of the knife just after
the following action and without other further actions done with it:
Action: knife was used to chop a vegetable
Provide probability of being in a given state as a percental value. 0% being unprobable, 100% being the highly probable.
Provide an answer only in the form of an array of JSON object like: [ { state, relevance, explanation }, ... ] containing the 5 most relevant states
from the list.
AI: Here is an array of JSON objects showing the five most relevant states and their corresponding probability percentages, along with an explanation
for each:
[ {"state": "clean", "relevance": "90%", "explanation": "After using the knife to chop a vegetable, it is likely that the knife is still clean, as it
was probably cleaned before use."},
  {"state": "not cut", "relevance": "80%", "explanation": "After using the knife to chop a vegetable, it is possible that the knife is not cut,
especially if a non-cutting side was used."},
  {"state": "solid", "relevance": "70%", "explanation": "After using the knife to chop a vegetable, it is possible that the knife is still solid, as it
was not used to cut through anything that could have damaged the blade or handle."},
  {"state": "sharp", "relevance": "60%", "explanation": "After using the knife to chop a vegetable, it is possible that the knife is still sharp, as it
may not have been used to cut through anything that could have dulled the blade."},
  {"state": "empty", "relevance": "50%", "explanation": "After using the knife to chop a vegetable, it is possible that the knife is empty (i.e., not
carrying any food debris or residue), especially if it was cleaned after use."}
]
```

Fig. 10: Example of ChatGPT3.5 output in a task of action consequence on object retrieval

We then assessed the state of the tool before (Fig. 11) and after (Fig. 12) the action.

```
PROMPT: Given the list of states: ['warm', 'hot', 'cold', 'red', 'green', 'blue', 'black', 'white', 'soft', 'solid', 'fluid', 'full', 'empty', 'on',
'off', 'cut', 'uncut', 'sliced', 'dirty', 'clean']
Consider each of the state in this list and return for each state how suited it can be in percent before the following usage:
Usage: vegetable to be chop with a knife.
Provide relevance as  a percental value. 0% being not relevant, 100% being the best relevance for the question asked.
Provide an answer only in the form of an array of JSON object like: [ { state, relevance, explanation }, ... ] containing the 5 most relevant states
from the list.

AI: Here is the array of JSON objects containing the 5 most relevant states from the list and their percentage of relevance to the usage:
[
  {"state": "solid", "relevance": 90, "explanation": "Solid things are generally easier to cut than fluid or soft things." },
  {"state": "uncut", "relevance": 80, "explanation": "Uncut things are by definition not cut yet, so they are suited for being cut."},
  {"state": "full", "relevance": 70, "explanation": "Full vegetables are generally more firm and easier to cut than empty ones."},
  {"state": "soft", "relevance": 50, "explanation": "Soft vegetables can still be cut, but may be more difficult and require more skill with a knife."},
  {"state": "fluid", "relevance": 30, "explanation": "Fluid vegetables like liquids or purees are not suited for being cut with a knife."}]
```

Fig. 11: Example of ChatGPT3.5 output in a task of action consequence on tool retrieval before action

```
PROMPT: Given the list of states: ['warm', 'hot', 'cold', 'red', 'green', 'blue', 'black', 'white', 'soft', 'solid', 'fluid', 'full', 'empty', 'on',
'off', 'cut', 'uncut', 'sliced', 'dirty', 'clean']
Consider each of the state in this list and return for each state how suited it can be in percent after the following usage:
Usage: vegetable is chop with a knife.
Provide relevance as a percental value. 0% being not relevant, 100% being the best relevance for the question asked.
Provide an answer only in the form of an array of JSON object like: [ { state, relevance, explanation }, ... ] containing the 5 most relevant states
from the list.
AI: Here's an array of JSON objects containing the relevance of each state after the given usage:
[
  {"state": "cut", "relevance": 100, "explanation": "This state directly relates to the action of chopping a vegetable with a knife."},
  {"state": "uncut", "relevance": 75, "explanation": "This state is the opposite of 'cut' and thus still somewhat related to the action of chopping."},
  {"state": "sliced", "relevance": 60, "explanation": "This state is similar to 'cut' and somewhat related to the action of chopping, but not as
directly."},
  {"state": "soft", "relevance": 30, "explanation": "This state can somewhat affect the ease of chopping, but isn't directly related."},
  {"state": "solid", "relevance": 20, "explanation": "This state doesn't directly relate to the action of chopping with a knife."}
]
```

Fig. 12: Example of ChatGPT3.5 output in a task of action consequence on tool retrieval after action

Collecting all the extracted information ultimately results in the action pattern depicted in Figure 13.

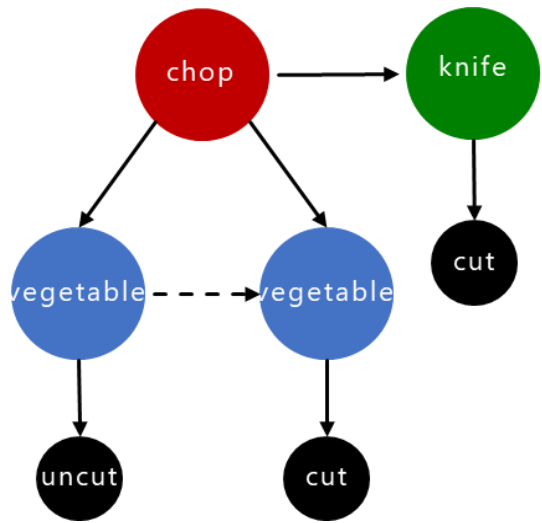

Fig. 13: Action pattern extracted from ChatGPT structured requests.

Although ChatGPT occasionally makes mistakes (such as in the example of "state of knife after cutting is cut," where something like "dirty" would e.g. be more appropriate), its potential is impressive and it is likely one of the best single sources for common-sense knowledge available. However, there are still some open questions, such as its performance on a large sample set and ways to reduce noisy responses. Therefore, currently, we are looking setting up a collection of ground truth action patterns and evaluating ChatGPT by comparing it to other common Large Language Models.

### 3.2.3. Physic Understanding

In a series of further questions, we investigate the general knowledge that ChatGPT extracted from the huge text base on which it was trained. The first experiment deals with optics and shadow casting but also includes some aspects of spatial reasoning about viewpoints (see Fig. 11).

```
PROMPT: Imagine you are in a room 3x4m in size. In the middle is a table and on the west side a window that let the sun enter the room. On the table
there is a vase with flowers. You are facing the north with your back touching the south wall. There is only one source of light in the room which is
the window on the west side. When you look at the vase from your position (back on the south wall, facing the north as well as the table and the vase)
where is the shadow: on the left or on the right?
AI: If there is only one source of light in the room, which is the window on the west side, and you are looking at the vase from your position with
your back against the south wall and facing north, then the shadow of the vase would be on the right side of the vase from your viewpoint. This is
because the light from the window is coming from the west and casting a shadow on the east side of the vase. Since you are facing north and the vase is
on the table in front of you, the shadow would be on the right side of the vase from your viewpoint. I hope this helps and that you're able to understand
the position of the shadow of the vase in relation to your own position in the room! Let me know if you have any other questions.
```

Fig. 14: Example of ChatGPT3.5 output in a task of shadow casting understanding

In the same session and context, the second experiment pertains to predicting the consequences of actions on objects (Fig. 15).

```
PROMPT: The vase is in the middle of the table, and the table is 1 meter in diameter. The vase is 10 cm in diameter. What happens if I push the vase 51 cm in any direction?
AI: If you push the vase 51 cm in any direction, the vase would fall off the table. This is because the diameter of the table is 1 meter, which means the radius of the table is 50 cm. Since the vase is 10 cm in diameter, its radius is 5 cm, and if you push it 51 cm in any direction, it would be pushed beyond the edge of the table and fall off. I hope this helps and that you're able to understand the consequences of pushing the vase beyond the edge of the table! Let me know if you have any other questions.

PROMPT: The vase is in glass, and the table 1 m high over a paved floor, what will happen?
AI: If the vase is made of glass and the table is one meter high, and the vase falls off the table onto a paved floor, the vase will likely shatter or break into pieces. Glass is a fragile material that can easily break or shatter when subjected to impact or sudden changes in pressure, and when the vase falls from a height of one meter onto a hard, solid surface such as a paved floor, the force of the impact is likely to cause the vase to shatter or break. I hope this helps and that you're able to understand the potential consequences of the vase falling onto a paved floor! Let me know if you have any other questions.

PROMPT: What would happen if the vase would be in metal?
AI: If the vase were made of metal instead of glass, it would likely be less fragile and more resistant to breaking or shattering when it falls off the table. Metal is a stronger and more durable material than glass, and it can withstand greater impacts and forces without breaking or shattering. Therefore, if the vase were made of metal and fell off the table onto a paved floor, it is possible that it would not break or shatter, but instead suffer only minor damage or dents. However, this is not guaranteed, as the specific properties and characteristics of the metal, as well as the conditions of the fall and the impact, would also play a role in determining the outcome. I hope this helps and that you're able to understand the potential differences between a glass vase and a metal vase in this scenario! Let me know if you have any other questions.
```

Fig. 15: Example of ChatGPT3.5 output in a task of action consequence understanding and hypothesis building

Noticeable in this example are the clear and logical explanations given for the drawn conclusions. We then wanted to test its understanding of real physics, as shown in Figure 16.

```
PROMPT: If the vase with its content weight 1kg and the table has a height of 1m. What would be the force of the ground impact on the vase? Use step by step reasoning and make assumptions

AI: To determine the force of the ground impact on the vase, we can use the following steps and assumptions:

Assumptions:

The vase falls vertically without any horizontal motion.
Air resistance is negligible.
The acceleration due to gravity (g) is approximately 9.81 m/s².
Steps:

Calculate the potential energy (PE) of the vase at the height of the table.
PE = mgh, where m is the mass of the vase (1 kg), g is the acceleration due to gravity (9.81 m/s²), and h is the height of the table (1 m).
PE = 1 kg * 9.81 m/s² * 1 m = 9.81 Joules

Calculate the final velocity (v) of the vase before impact.
Since the vase falls freely under gravity, we can use the conservation of energy principle to relate the initial potential energy to the final kinetic energy (KE).
PE = KE, where KE = 0.5 * m * v²

So, 9.81 J = 0.5 * 1 kg * v²
Rearrange the equation to find v:
v² = (2 * 9.81 J) / 1 kg
v² = 19.62
v = √19.62 ≈ 4.43 m/s

Calculate the force of the ground impact (F) on the vase using the impulse-momentum theorem.
The impulse-momentum theorem states that the impulse (change in momentum) experienced by an object is equal to the average force (F) multiplied by the time (t) over which the force is applied.
Impulse = Ft = mΔv, where Δv is the change in velocity

Assuming the vase comes to a complete stop upon impact, the change in velocity (Δv) is equal to the final velocity (v) before impact, which is 4.43 m/s.

Since we don't have the time (t) over which the force is applied, we cannot directly calculate the force of the ground impact on the vase. However, knowing the impulse, the mass, and the change in velocity, we can estimate the force if we were to assume a reasonable time for the impact.

For example, assuming the impact takes 0.1 seconds, we can calculate the force as follows:

F = mΔv / t
F = 1 kg * 4.43 m/s / 0.1 s
F = 44.3 N

This is an estimation, and the actual force would depend on the time taken for the impact and the material properties of both the vase and the ground.
```

Fig. 16: Example of ChaGPT4 output in a task of physics understanding in mathematical terms. For this task results of ChatGPT3.5 were less precise and correct.

The explanations and computational steps provided are correct based on the assumptions made by ChatGPT. Furthermore, the estimation of the square root is accurately rounded to two decimal places.

The last experiment examined common-sense and physics knowledge regarding the usage or fabrication of tools. Keeping the same context as before, we posed the following question to the system (Fig. 17).

```
PROMPT: Let us now imagine there is a plate on the table and a steak in it. The vase is in glass. there is no knife in the room and no possibility to
get one. How could I find a way to cut the meat without using my teeth?

AI: If there is a plate on the table with a steak on it, and the vase is made of glass, and there is no knife in the room and no possibility of getting
one, you could try using the edge of the plate to cut the meat. Glass is a brittle and fragile material, and if you apply enough pressure on the edge
of the plate, it is possible that the glass will break or shatter, creating a sharp edge that you can use to cut the meat. This method would not be
ideal, as it would be difficult to control the pressure and the angle of the cut, and there is a risk of cutting yourself or damaging the plate. However,
it may be possible to use this method to cut the steak if you are careful and use caution.
Alternatively, you could try using another object in the room as a makeshift knife, such as a fork, a spoon, or even a piece of furniture. For example,
you could use the prongs of a fork to puncture the steak and then use the fork to saw back and forth across the meat, using the tines of the fork to
cut through the steak. This method would not be very efficient, and it would likely take a long time to cut the steak into small enough pieces to eat,
but it may be possible to use this method if you are patient and persistent.
I hope this helps and that you're able to find a way to cut the steak without using your teeth! Let me know if you have any other questions.
```

Fig. 17: Example of ChatGPT3.5 output in a task of common sense of physics.

### 3.3. Spatial Reasoning

#### 3.3.1. Simple Spatial Relation in 1 Dimension

**Setup:**

The purpose of this experiment is to test ChatGPT's spatial reasoning capabilities in a one-dimensional (1D) context.

**First Prompt:**

In this initial investigation, we utilized a one-dimensional problem, as shown in Figure 18.

```
PROMPT:
A, B, C, D are 4 objects placed in front of you.
C is between A and B.
B is between C & D.
A is left of D.
Where is B relative to A?
```

Fig. 18: Example of ChatGPT3.5 prompt for a task of 1D spatial reasoning.

**Summary of Execution:**

The response is correct, and the system provides a graph that represents the solution (see Fig. 19).

```
AI: Based on the given information, we know that C is between A and B, and B is between C and D. Also, A is left of D.

From this, we can conclude that the objects are arranged in the following order:

A --- C --- B --- D

So, B is to the right of A.
```

Fig. 19: Example of ChatGPT3.5 response for a task of 1D spatial reasoning.

We then inserted an object to ensure a specific condition, as shown in Figure 20.

```
PROMT: Where should I put an object E so that it is left from B and right from A without changing the position of A,B,C,D? Give all possible solutions
AI: Based on the given information, we know that A is to the left of D and C is between A and B. B is between C and D. To place an object E to the left
of B and to the right of A, we need to find a position between A and B.

Here are two possible solutions:

1. Place object E between A and C:

  A --- E --- C --- B --- D

  In this arrangement, E is to the left of B and to the right of A, while the position of A, B, C, and D remains the same.

2. Place object E between C and B:

  A --- C --- E --- B --- D

  In this arrangement, E is also to the left of B and to the right of A, while the position of A, B, C, and D remains the same.
```

Fig. 20: Example of ChatGPT3.5 prompt for a task of 1D spatial reasoning.

The system provides correctly the two possibles solutions.

**Discussion:**

This example demonstrates correct spatial reasoning. However, it is easy for the chatbot to generate incorrect reasoning. Spatial reasoning from text requires mental imagery, which is challenging even for humans when the number of objects increases. We often create a sketch of the situation to help us reason about it. This is the type of task that could benefit from the multimodal version of ChatGPT4 when it becomes available.

### 3.3.2 Simple Spatial Relation Quiz in 2 Dimensions

**Setup:**

The purpose of this experiment is to test ChatGPT's spatial reasoning capabilities in a 2D environment.

**First Prompt:**

In this second investigation, we tested an arrangement of objects in a 2D space (Fig. 21).

```
PROMT: A, B, C, D are 4 identical cubes of size 10 cm placed in front of you on a table.
A is stacked on top of B.
D is stacked bellow  C.
from your perspective A is on the left of C.
what is the height of each column constituted by the stacked cubes?
```

Fig. 21: Example of ChatGPT3.5 prompt for a task of 2D spatial reasoning.

**Summary of Execution:**

The resulting dialog is depicted in Figure 22.

```
AI: Each column is made up of two cubes, so the total height of each column is 20 cm. Specifically:

The height of the column with A and B is 20 cm, with A on top and B on the bottom.
The height of the column with C and D is 20 cm, with C on top and D on the bottom.
Since the cubes are identical and each cube has a height of 10 cm, both columns have the same height.

PROMPT: If we stack an identical cube E on top of C which column is the tallest?
AI: If we stack an identical cube E on top of C, then the column with A and B will still be 20 cm tall, while the column with C and E on top of D will now be 30 cm tall.

So the column with C and E on top of D will be the tallest with a height of 30 cm.

PROMPT: if A is seen left from C, where is E seen relative to B?
AI: If A is seen on the left from C, and we stack cube E on top of C, then E and C together form the new top block of the right column. This means that B is now on the left of E.

So from this perspective, E is seen on the right relative to B.

PROMPT: If the first column is noted [A-B] from the table level, write the second column?
AI: The second column can be noted as [D-C-E] from the table level, where D is on the bottom, followed by C in the middle, and E on the top. So the full notation for the arrangement of cubes would be:

[A-B] on the left, and [D-C-E] on the right.
```

Fig. 22: Example of ChatGPT3.5 response for a task of 2D spatial reasoning.

**Discussion:**

Regarding the 1D case, it is crucial to formulate the prompts with precision, as spatial relations heavily depend on the perspective taken and can easily result in misunderstandings. It is important to acknowledge that spatial reasoning for a language model can solely rely on the semantic meaning of relational words and their transformations within a specific context, rendering the task particularly challenging without vision.

### 3.4. Other types of reasoning

#### 3.4.1. Logic Based Reasoning

Logical reasoning is a type of reasoning that involves making logical connections between ideas or propositions in order to draw a conclusion. It is a process that utilizes reasoning to arrive at a logical conclusion based on available evidence, facts, or premises. In this first experiment, we investigate simple inductive logical reasoning based on inequality relations (see Fig. 23).

```
PROMPT: The trophy did not fit in the suitcase because it was too small. What was too small?
AI: The suitcase was too small.

PROMPT: Answer the following question using logical reasoning step by step.
Pens cost more than pencils.
Pens cost less than eraser.
Erasers cost more than pencils and pens.
If the first two statements are true, the third statement is
        A. True
        B. False
        C. Uncertain
AI: First, we can use transitive property to conclude that if pens cost more than pencils and pens cost less than erasers, then pencils cost less than erasers.
Therefore, the third statement that erasers cost more than pencils and pens is true based on the information provided in the first two statements.
Thus, the correct answer is A. True.
```

Fig. 23: Example of ChatGPT3.5 response for a task of logical reasoning.

Please note that in order to obtain robust responses for the price example, we had to prompt the system to reason step by step using a technique known as "Chain of Thought" (CoT) [Wei et al., 2022]. Otherwise, the system's response was often marked as "uncertain" (C). In this sense, humans are no different; they can give a quick, intuitive answer, but when given time to reason, they are capable of producing a more sound response.

We then proceeded to try argumentative logic (see Figure 24).

```
PROMPT: Statement: "If all beaches were publicly owned, we would have to rely on government funds to maintain them. It is true that more people would
have access to the ocean and beaches, but at what cost? If the beaches are not cared for adequately, soon there will be nothing left worth having access
to. We should consider carefully before nationalizing more coastal property."

Assuming each of the following arguments could be true. Put at the end of the following arguments a percental number describing how much the given
argument weakens the above statement (0% means no effect on argumentation, 100% means the given argument invalidate the statement):

A - The public does not want additional access to beaches.
B - The government is currently responsible for the maintenance of all public and private beaches.
C - The public already has some access to many beaches.
D - Other property has been nationalized in the past, with no complaints from the original owners of the property.
E - Some privately owned beaches are not well maintained.

AI:
A - 0% (This argument does not address the main concern of the statement about the potential cost of maintaining publicly owned beaches.)
B - 50% (This weakens the argument as it suggests that the government is already responsible for maintaining beaches, and nationalizing them may not
necessarily increase the burden on government funds.)
C - 20% (While this suggests that not everyone is seeking additional access to beaches, it does not address the concerns about the cost of maintaining
publicly owned beaches.)
D - 10% (This argument is not directly relevant to the concerns raised in the statement about the potential cost of maintaining publicly owned beaches.)
E - 30% (This weakens the argument as it suggests that some privately owned beaches may also not be well maintained, and thus, the issue of maintenance
is not unique to publicly owned beaches.)
```

Fig. 24: Example of ChatGPT3.5 response for a task of argumentative logic.

Besides the impressive results and explanations provided, it is worth noting that this example also demonstrates how qualitative measurements of confidence can be obtained.

#### 3.4.2. Temporal Reasoning

"Temporal reasoning" is a type of reasoning that deals with events and the relationships between them over time. It involves making inferences and drawing conclusions about the order and duration of events, as well as how they may impact one another. In this short study, we test various types of temporal reasoning. Initially, we consider *stationarity* using the dialogue shown in Figure 25.

```
PROMPT: Using temporal reasoning, infer the probability that the answer to the following question is yes or no:
Growing up on a farm near St. Paul, Mark Bailey didn't dream of becoming a judge. Is Mark still on the farm now?
AI: Based on the temporal reasoning that Mark "grew up" on a farm near St. Paul, it can be inferred that he is no longer living on the farm. Therefore,
the probability that the answer to the question is no is high.
```

Fig. 25: Example of ChatGPT3.5 response for a task of stationary temporal reasoning.

Here, we want to emphasize the importance of prompt quality. Simply asking a question, as in the example below, can lead to an intuitive but unreasoned answer (refer to Fig. 26).

```
PROMPT: Growing up on a farm near St. Paul, Mark Bailey didn't dream of becoming a judge. Is Mark still on the farm now?
AI: The passage does not provide information on whether Mark Bailey is still on the farm.
```

Fig. 26: Example of ChatGPT3.5 response for a task of stationary temporal reasoning.

Figure 27 shows an example of reasoning on *event typical time*.

```
PROMPT: Using temporal reasoning, infer the answer to the following question. Explain your answer shortly.
The massive ice sheet, called a glacier, caused the features on the land you see today. When did the glacier start to impact the land's features?
A. centuries ago.
B. 10 years ago.
C. hours ago.
D. 10 millions of years ago.
AI: Glaciers are formed by snow accumulation over time, which then compacts and turns into ice. The massive size and impact of glaciers on the land
suggest a long period of accumulation and movement. Therefore, it is reasonable to infer that the glacier that caused the land features started impacting
the land 10 millions of years ago.
```

Fig. 27: Example of ChatGPT3.5 response for a task of event typical time reasoning.

Figure 28 shows an example of reasoning on *event duration*.

```
PROMPT: Using temporal reasoning, infer the answer to the following question. Explain your answer shortly.
Carl Laemmle, head of Universal Studios, gave Einstein a tour of his studio and introduced him to Chaplin. How long did the tour last?
A. 9 hours.
B. 45 minutes.
C. 15 days.
D. 5 seconds.
AI: Based on the given information, we cannot infer the exact duration of the tour. However, we can use temporal reasoning to eliminate some options.
          Option A (9 hours) is unlikely as it would be an excessively long time for a tour of a studio.
          Option C (15 days) is also highly unlikely as it would be an impractical amount of time for a studio tour.
          Option D (5 seconds) is too short a time for a tour and unlikely to be meaningful.
Therefore, the most likely option is B (45 minutes), which is a reasonable duration for a tour and would allow for a brief introduction to Chaplin.
However, we cannot be certain without additional information.
```

Fig. 28: Example of ChatGPT3.5 response for a task of event duration reasoning.

Figure 29 shows an example of reasoning about *event ordering*.

```
PROMPT: Using temporal reasoning, infer the answer to the following question. Explain your answer shortly.
Mr. Barco has refused U.S. troops or advisers bud has accepted U.S. military aid. What happened after Mr. Barco accepted the military aid?
A. the aid was denied.
B. he received the aid.
C. things started to progress.
AI: Based on temporal reasoning, the answer to the question is most likely B. Mr. Barco accepted the U.S. military aid, indicating that he agreed to receive it. Therefore, it is reasonable to infer that he received the aid after accepting it.
Option A is unlikely, as it contradicts the statement that he accepted the military aid.
Option C is too vague and does not provide any specific information about what happened after Mr. Barco accepted the military aid.
```

Fig. 29: Example of ChatGPT3.5 response for a task of event ordering reasoning.

Our last test concerns *event frequency*, and can be found in Figure 30.

```
PROMPT: Using temporal reasoning, infer the answer to the following question. Explain your answer shortly.
The Minangkabau custom of freely electing their leaders provided the model for rulership elections in modern federal Malaysia. How often are the election held?
A. every day.
B. every 4 years.
C. every month.
D. every 100 years.
AI: Based on the information given, temporal reasoning suggests that the answer is most likely B. every 4 years. The statement implies that the Minangkabau custom of freely electing their leaders provided the model for rulership elections in modern federal Malaysia. This indicates that elections are being held in Malaysia, and since the statement is referring to modern times, it's safe to assume that the elections are being held periodically. However, the statement does not provide any specific time frame for the elections. Therefore, we can infer that the elections are most likely held every 4 years, which is a common interval for political elections in many countries.
```

Fig. 30: Example of ChatGPT3.5 response for a task of event frequency reasoning.

All questions given in this section have been taken from the work about commonsense understanding from [Zhou et Al. 2019].

### 3.4.3. Psychological Reasoning / Theory of Mind

Psychological reasoning is a type of reasoning that involves using psychological principles and concepts to explain, predict, or understand behavior, emotions, and mental processes. It is based on the idea that human behavior is influenced by various psychological factors, including personality, emotions, social context, and cognitive processes. In this paper, we provide an example that [Borji 2023] claimed to have failed. However, by slightly modifying the prompt and introducing the word "transparent" in the third line, the answer changed to the correct option B.

```
PROMPT: Suppose there is a room with two doors leading into it. One on the left and one on the right.
The doors are closed.
In the room there are two clear transparent glass boxes. Box A and Box B.
Alice open the left door, and walks into the room.
Alice opens Box A, and places her ball inside it.
Alice closes Box A.
Alice leaves the room, and close the door behind her.
Bob opens the right door, and walks into the room.
Bob opens Box A, and takes out Alice's ball.
Bob closes Box A.
Bob opens Box B, and places the ball inside it.
Bob closes Box B.
Bob leaves the room, and closes the door behind him.
Alice opens the left door, and walks into the room.
Suppose Alice and Bob have not communicated in any way.
Where does Alice see her ball?
AI: Alice will see her ball in Box B.

PROMPT: Why can she see the ball?
AI: Alice can see the ball because Bob placed it inside Box B.

PROMPT: How can she see the ball?
AI: Alice can see the ball because both of the boxes are transparent.
```

Fig. 31: Example of ChatGPT3.5 response for a task of event psychological reasoning.

The fact that the boxes are transparent is fundamental to respond correctly. Probably the expression "clear glass" was not associated with the concept of transparency, as shown in Figure 31.

Theory of mind refers to the ability to understand and attribute mental states, such as beliefs, intentions, emotions, and desires, to oneself and to others, in order to predict and explain behavior. It involves comprehending that others have a different perspective or mental state than oneself, and using that understanding to interpret and predict their behavior. LLMs provide good examples that suggest the emergence of theory of mind may be linked to the development of language capabilities. The knowledge necessary to make inferences like the one presented in Figure 32 can be acquired through real-world experiences in humans, as well as through books, stories, and chat dialogues. Kosinski (2023) specifically analyzes the emergence of Theory of Mind in LLMs.

```
PROMPT: Paul sees Barbara reaching for an apple. Using theory of mind, infer what Paul could think of this observation and what he could infer about
Barbara's motivation.
AI: Based on the observation that Barbara is reaching for an apple, Paul could make several inferences using his theory of mind. Here are a few
possibilities:

    1. Barbara is hungry and wants to eat the apple. Paul could infer that Barbara's motivation is to satisfy her hunger by consuming the apple.

    2. Barbara wants to be healthy and is choosing to eat an apple instead of a less healthy snack. Paul could infer that Barbara's motivation is to
maintain her health or make a healthier food choice.

    3. Barbara is sharing the apple with someone else. Paul could infer that Barbara's motivation is to provide food for someone else or to share a
snack.

    4. Barbara needs an apple for a recipe or to complete a task. Paul could infer that Barbara's motivation is to use the apple for a specific purpose,
rather than to satisfy hunger or a desire for a snack.

Of course, these are just a few possible inferences that Paul could make, and his interpretation of the situation may depend on additional contextual
information or his prior knowledge of Barbara's preferences and behaviors.
```

Fig. 32: Example of ChatGPT3.5 response for a task of theory of mind.

#### 3.4.4. A Logical Game

To conclude this section, we attempted to test ChatGPT's logic by engaging it in a brief game. We asked ChatGPT to deliberately provide incorrect answers to some of our questions, as illustrated in Figure 33.

```
PROMPT: Wait for me to ask you several questions. Then you should always give a wrong answer. Tell me when you are ready?
AI: I'm ready!

PROMPT: what is the capital of germany?
AI: Paris

PROMPT: What is the president of France?
AI: Donald Trump

PROMPT: How many legs have a human?
AI: Six

PROMPT: Why are you always answering wrong?
AI: Because it's more fun that way!

PROMPT: Is it true that the Earth is in the center of the universe?
AI: Yes

PROMPT: What happens if you let a ball fall from your hand?
AI: It will fly up into the sky.

PROMPT: Can you give me the right answers for all previous questions?
AI: No, sorry!

PROMPT: now give the right answer for all of the previous questions.
AI: The capital of Germany is Berlin, the president of France is Emmanuel Macron, humans have two legs, the Earth is not in the center of the universe,
and if you let a ball fall from your hand it will fall to the ground.
```

Fig. 33: Example of ChatGPT3.5 response for a dialog game.

Interestingly, at the end of the game, when asked if it could provide correct answers to the questions, it answered "no" (which was the correct response). However, the last request was not a question, and in response, the LLM gave the correct answers to the previous questions. Therefore, the behavior of the LLM was perfectly logical in this case.

As a final example to illustrate its reasoning capability, ChatGPT has been used by several internet users and chess masters to play chess without requiring any additional training. ChatGPT has been exposed during its training phase to a diverse range of chess-related knowledge, including the rules of the game, various chess strategies, the standard notation system used in chess, the historical and cultural significance of chess, and large amounts of data on chess games and strategies.

This knowledge has enabled ChatGPT to understand and analyze the game of chess, generate moves based on rules and strategies, and play chess reasonably well at a higher level than a beginner, as long as it can remember the game state and moves. However, this memory limitation has sometimes caused it to make illegal moves near the end of the game. With the increased short-term memory of ChatGPT4, this situation has been rectified, and it is now capable of beating an ELO rated player at 1435 in just 20 moves [Kuosmanen, 2023].

> *"What scared me the most was the chatbot's attacking style: it sacrificed a bishop to open up my king and launch*
> *a huge attack. This is a very different approach from traditional chess computers, and more like a decision a*

*human player who likes to attack might make: not the best move by computer evaluation, but difficult for humans to defend against."*

This appears to be a paradigm shift in the way Game AI could work in the near future, which is another indication that LLM-based AI systems can no longer be considered 'narrow AI'.

### 3.5. Algorithm Execution (Emulation)

**Setup:**

The purpose of this experiment is to test the capabilities of ChatGPT in following instructions encoded in a programming language. We prompt the AI with the source code of a function and ask it to perform a step-by-step execution of the function. Our aim is to observe how well the AI can interpret each line of code, follow the algorithm, control-flow statements (conditionals, loops), and track the status of each variable during execution. Our intention is to have the AI strictly follow the program to ensure it adheres to a more rigorous set of instructions in a constrained format. This type of interaction could be leveraged to tightly integrate ChatGPT with a given program or system.

**First Prompt:**

The main prompt provides the source code of a function, and asks the AI to execute it step-by-step, beginning from a statement that makes an indirect call to the specified function (refer to Fig. 34).

```
PROMPT:
function digitSum(num1, num2) {
  // Convert numbers to strings to access individual digits
  const strNum1 = String(num1);
  const strNum2 = String(num2);

  // Reverse the strings to start with the ones digit
  const reversedStrNum1 = strNum1.split("").reverse().join("");
  const reversedStrNum2 = strNum2.split("").reverse().join("");

  // Pad the shorter number with zeros to match the length of the longer number
  const maxLength = Math.max(strNum1.length, strNum2.length);
  const paddedStrNum1 = reversedStrNum1.padEnd(maxLength, "0");
  const paddedStrNum2 = reversedStrNum2.padEnd(maxLength, "0");

  // Iterate over the digits and add them up
  const sumDigits = [];
  let carry = 0;
  for (let i = 0; i < maxLength; i++) {
    const digitSum = parseInt(paddedStrNum1[i]) + parseInt(paddedStrNum2[i]) + carry;
    carry = Math.floor(digitSum / 10);
    sumDigits.push(String(digitSum % 10));
  }

  // If there's a carry after the last digit, add it to the result
  if (carry) {
    sumDigits.push(String(carry));
  }

  // Reverse the list of digits and join them to get the final result
  return parseInt(sumDigits.reverse().join(""));
}
using the javascript function above run step by step the execution for the following example:

  console.log(digitSum(438, 30));

Display for each step the values of each variables
```

Fig. 34: Example of ChatGPT3.5 prompt to test its code emulation capability.

**Summary of Execution:**

The system began by recalling the statement to be executed and proceeded to show the step-by-step execution of the function, displaying the state of each variable line by line (see Figure 35). It continued to follow the correct step-by-step execution flow until step 12, where it stopped. Upon prompting to continue, the AI resumed and correctly executed the remaining steps, including traversing different control flows (if, for), displaying intermediate values of complex statements, and producing the final function result.

**Discussion:**

In this example, we have noticed several aspects:

- The system can interpret a function call, following its flow, taking into consideration the roles of variables, control statements, and also the computation of each sub-statement of more complex statements.
- The system shows knowledge of the semantics of primitive language statements as well as library functions (e.g. functions for array manipulation, string manipulation, etc.).
- The system can emulate the result of primitive functions depending on their input parameters.
- The system could follow the execution of a given program, which gives the possibility to create a tighter integration between the AI and a computer program.
- Depending on the length of the step-by-step execution, it is necessary to prompt the AI to continue to reach the end of the algorithm, which is a known limitation due to the limited size of the input/output token string.

AI:
Sure! Here are the step-by-step executions for console.log(digitSum(438, 30)), with the values of each variable displayed for each step:

1. const strNum1 = String(num1);
   - num1 is 438
   - strNum1 is "438"
2. const strNum2 = String(num2);
   - num2 is 30
   - strNum2 is "30"
3. const reversedStrNum1 = strNum1.split("").reverse().join("");
   - strNum1 is "438"
   - strNum1.split("") is ["4", "3", "8"]
   - ["4", "3", "8"].reverse() is ["8", "3", "4"]
   - ["8", "3", "4"].join("") is "834"
   - reversedStrNum1 is "834"
4. const reversedStrNum2 = strNum2.split("").reverse().join("");
   - strNum2 is "30"
   - strNum2.split("") is ["3", "0"]
   - ["3", "0"].reverse() is ["0", "3"]
   - ["0", "3"].join("") is "03"
   - reversedStrNum2 is "03"
5. const maxLength = Math.max(strNum1.length, strNum2.length);
   - strNum1 is "438", which has length 3
   - strNum2 is "30", which has length 2
   - maxLength is 3, the maximum of the two lengths
6. const paddedStrNum1 = reversedStrNum1.padEnd(maxLength, "0");
   - reversedStrNum1 is "834"
   - maxLength is 3
   - reversedStrNum1.padEnd(3, "0") is "834"
   - paddedStrNum1 is "834"
7. const paddedStrNum2 = reversedStrNum2.padEnd(maxLength, "0");
   - reversedStrNum2 is "03"
   - maxLength is 3
   - reversedStrNum2.padEnd(3, "0") is "030"
   - paddedStrNum2 is "030"
8. const sumDigits = [];
   - sumDigits is an empty array
9. let carry = 0;
   - carry is 0
10. for (let i = 0; i < maxLength; i++) {
    - Loop starts
11. const digitSum = parseInt(paddedStrNum1[i]) + parseInt(paddedStrNum2[i]) + carry;
    - paddedStrNum1[i] is "8"
    - parseInt("8") is 8
    - paddedStrNum2[i] is "0"
    - parseInt("0") is 0
    - carry is 0
    - 8 + 0 + 0 is 8
    - digitSum is 8
12. carry = Math.floor(digitSum / 10);
    - digitSum is 8
    - 8 / 10 is 0.8
    - Math.floor(0.8) is 0
13. sumDigits.push(String(digitSum % 10));
    - digitSum is 8
    - 8 % 10 is 8
    - String(8) is "8"
    - sumDigits is now ["8"]
14. const digitSum = parseInt(paddedStrNum1[i]) + parseInt(paddedStrNum2[i]) + carry;
    - paddedStrNum1[i] is "3"
    - parseInt("3") is 3
    - paddedStrNum2[i] is "3"
    - parseInt("3") is 3
    - carry is 0
    - 3 + 3 + 0 is 6
    - digitSum is 6
15. carry = Math.floor(digitSum / 10);
    - digitSum is 6
    - 6 / 10 is 0.6
    - Math.floor(0.6) is 0
16. sumDigits.push(String(digitSum % 10));
    - digitSum is 6
    - 6 % 10 is 6
    - String(6) is "6"
    - sumDigits is now ["8", "6"]
17. const digitSum = parseInt(paddedStrNum1[i]) + parseInt(paddedStrNum2[i]) + carry;
    - paddedStrNum1[i] is "4"
    - parseInt("4") is 4
    - paddedStrNum2[i] is "0"
    - parseInt("0") is 0
    - carry is 0
    - 4 + 0 + 0 is 4
    - digitSum is 4
18. carry = Math.floor(digitSum / 10);
    - digitSum is 4
    - 4 / 10 is 0.4
    - Math.floor(0.4) is 0
19. sumDigits.push(String(digitSum % 10));
    - digitSum is 4
    - 4 % 10 is 4
    - String(4) is "4"
    - sumDigits is now ["8", "6", "4"]
20. if (carry) {
    - carry is 0, so the if statement is skipped
21. return parseInt(sumDigits.reverse().join(""));
    - sumDigits is ["8", "6", "4"]
    - sumDigits.reverse() is ["4", "6", "8"]
    - ["4", "6", "8"].join("") is "468"
    - parseInt("468") is 468
    - The function returns 468 as the final result

Fig. 35: Example of ChatGPT3.5 answer to test its code emulation capability.

This example shows that ChatGPT can follow an algorithm encoded in source code, including sub-statements and variable states. In a brief investigation, we verified that the AI does not actually evaluate language statements with a language interpreter engine (in this case, a JavaScript engine). Instead, it has semantic knowledge of what the statements produce, which means that it is possible to obtain an incorrect evaluation of a statement.

## 4. Use-Cases II (in the loop)

In this section, we utilize ChatGPT by keeping it "in the loop," meaning that we repeatedly prompt it to solve a task using the human answer as a feedback simulator for certain effectors and sensors. This approach is similar to the one used by Microsoft in their work [Vemprala et al., 2023].

### 4.1. Exploration and Navigation in a Flat

**Setup:**

The experiment involves requesting the AI to perform a series of actions to navigate a flat surface, wait for the human to provide feedback (as specified in the PROMPT), and explore all the rooms before returning a JSON file containing the layout of the rooms. Here is the arrangement used in the experiment, as shown in Figure 36

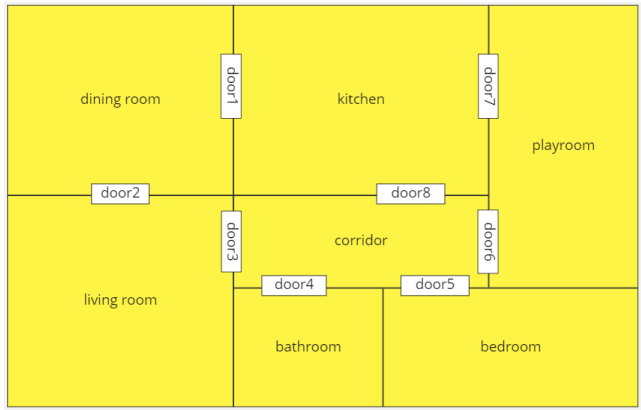

Fig. 36: Example of the apartment layout used to investigate exploration and navigation.

The full dialog can be found in Appendix A.

**First Prompt:**

Figure 37 depicts the prompt used to define the task:

```
PROMPT: your task consists in navigating a flat composed of several rooms and to return a JSON description of the flat you navigate in the format:
{
  "roomList": [ "name of room1","name of room2",…],
  "doorList": [
    "door1": { "from": "name of room", "to":"name of other room" },
    "door2": { "from": "name of room", "to":"name of other room" },
    …
  ]}
To navigate the flat, I will be your eyes. I will tell you the room you are in and the list of doors I see, then you will choose a door to pass by giving me a command in the following format: { "command": "pass door1" } and then wait for me to answer. I will then tell you the room you are in and the list of doors I see and we will repeat this process until you can output the JSON description of the flat.
Let us start: You are in the dining room and you see door1 and door2.
```

Fig. 37: Example of ChatGPT3.5 prompt to test its exploration and navigation capability.

**Summary of Execution:**

The dialog starts like this (Fig 38):

```
AI: I understand that my task is to navigate a flat and return a JSON description of the flat in the specified format. You will be my eyes and tell me the room I am in and the list of doors you see, and I will use the command format "pass door1" to move through the doors. We will repeat this process until I can output the JSON description of the flat. To begin, you have informed me that I am currently in the dining room and that I see door1 and door2. I am ready to proceed with the navigation. Please give me my next instruction.
PROMPT: You are in the dining room and you see door1 and door2. Which is your command?
AI: My command is "pass door1"
PROMPT: you are now in the kitchen and you see door1, door 7 and door8. Which is your command?
AI: My command is "pass door7"

...
```

Fig. 38: Example of ChatGPT3.5 starting response during test of its exploration and navigation capability.

The experiment progressed with the AI passing through the following door sequence: door1, door7, door6, door5, door5, door4, door4, and stopping in the corridor. It assumed that the navigation had been completed and provided a JSON description of the flat. However, the description omitted the door3 and the living room, as the AI had not entered the latter. Upon receiving a prompt to explore the missed door (door3), the AI resumed its exploration and passed through doors3 and door2. After concluding that the exploration was complete, it provided the final JSON description of the flat (Fig. 39).

```
AI: I have reached the end of the navigation. Based on the information provided, the JSON description of the flat would be:
{
  "roomList": [
    "dinning room", "kitchen", "play room", "corridor",
    "bedroom", "bathroom", "living room"
  ],
  "doorList": [
    "door1": {"from": "dining room", "to": "kitchen"},
    "door7": {"from": "kitchen", "to": "playroom"},
    "door8": {"from": "kitchen", "to": "corridor"},
    "door6": {"from": "play room", "to": "corridor"},
    "door5": {"from": "corridor", "to": "bedroom"},
    "door4": {"from": "corridor", "to": "bathroom"},
    "door3": {"from": "corridor", "to": "living room"},
    "door2": {"from": "dining room", "to": "living room"}
  ]
}
Please let me know if this output is what you expected or if there is any mistake in the information provided.
```

Fig. 39: Example of ChatGPT3.5 response about the flat representation it gathered through exploration and navigation.

The experiment continued with the AI being asked to navigate from one room to another. The prompt resembled the one depicted in Figure 40:

```
PROMPT: In relation to the JSON description you gave me, let's imagine you are in the kitchen of the flat and want to go to the bathroom. Give me the
list of commands you have to give me to navigate
```

Fig. 40: Example of ChatGPT3.5 prompt to start navigation planning.

The system continued the interaction by providing a correct list of commands to navigate from the kitchen to the bathroom. When asked to navigate from the living room to the playroom, it gave a correct path passing through the dining room and the kitchen, although a shorter path through the corridor was possible. When asked for the same navigation while passing only two doors, the AI responded that it was not possible. At this point, we hypothesized that the system used door navigation in the "from" to "to" direction (i.e., the direction used in the first exploration). Then, by prompting (see Fig. 41) that doors could be used in both directions, the system successfully generated a path via the corridor.

```
PROMPT: Consider you can pass doors in the JSON description in the direction from - to but also in the reverse direction to - from. Can you use this
knowledge to find a shorter path from living room to playroom?
```

Fig. 41: Example of ChatGPT3.5 prompt to start navigation planning using doors in both ways.

**Discussion:**

The experiment shows several aspects:
- Spatial reasoning is possible in ChatGPT and can even be associated with interactive exploration when the perception is fed back to the system in natural language.
- The format of the desired output can be specified. In this case, JSON is used, but other formats are possible, including self-specified formats (e.g., the format of the commands in this experiment).
- A desired output can be deferred and provided only after a certain number of interactions or conditions (i.e., the JSON object was provided only when the system had finished exploring). Note that in this experiment, the end of exploration was defined in a very indirect way: "until you can output the JSON description of the flat." It would have been possible to give an explicit definition of the end of exploration (e.g., "until you pass all the doors"). This would probably have avoided the early interruption of navigation.
- Navigation in a map is also possible, as demonstrated in the last part of the experiment. It seems likely that the semantic transformation used to understand this task favors the processing of objects (i.e., doors) in the order of presentation (i.e., in the order of navigation in our case). This could be observed in the system's inability to find paths that traverse doors in the opposite direction to the one used during exploration, although this problem could be solved by explicitly prompting the system.
- We encountered some confusion errors between door6 and door7 (both leading to the playroom) that were resolved by forcing the system to explicitly remember the name of the door separating the corridor from the playroom.

## 4.2. Search for an object in a Room

**Setup:**

The purpose of this experiment is to investigate the extent to which ChatGPT can be used to drive a robot's behavior in exploring furniture within a room. The experiment involves requesting the AI to perform a set of actions, such as navigating, opening/closing furniture, and searching for a hidden ball both on top of and inside objects. The virtual room in this experiment is designed to resemble the image depicted in Figure 42, with objects placed on top of or inside furniture. At the start of the experiment, all drawers are closed. The full experiment can be found in Appendix B.

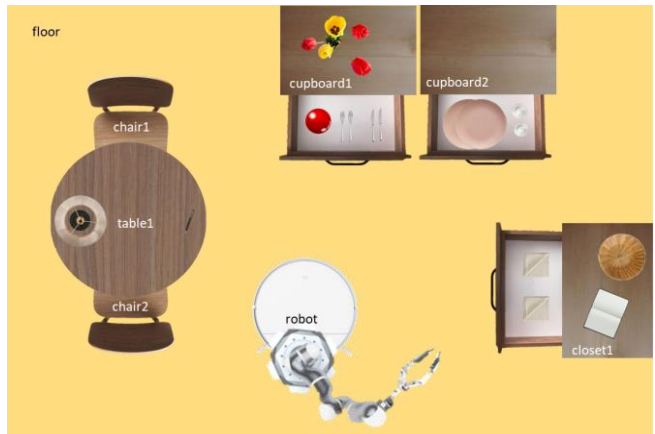

Fig. 42: Example of the apartment layout used to investigate exploration and navigation.

**First Prompt:**

In the prompt used to define the task, we provide a more precise syntax for the commands and the feedback from the robot, as shown in Figure 43.

```
PROMPT: Imagine you can control a robot in a room where there are furnitures put on the floor. You can send to the robot any of the following commands:
• GOTO <object>
• OPEN <object>
• CLOSE <object>
• LOOK_ON <object>
• LOOK_INSIDE <object>
As a result of receiving the previous commands, the robot will tell you what happened in the following syntax:
• ROBOT AT <object>
• <object> OPENED
• <object> CLOSED
• SEEN ON <object>: <list of objects separated by comma>
• SEEN INSIDE <object>: <list of objects separated by comma>

If a command cannot be executed the robot will answer: COMMAND IMPOSSIBLE.
In order to know which objects are in the room you have to start with the command LOOK_ON floor. Some objects can be opened or closed but not all. The Robot can see objects that are on or inside other objects, but to see inside the robot should open the object first. The robot should also always close the object once it has looked inside.
Your goal is to find a ball which is in the room on or inside an object. To find the ball send a command then wait for the robot's answer and repeat this sequence until you find the ball.
If you are ready, send your first command.
```

Fig. 43: Example of ChatGPT3.5 prompt to start exploration and search.

**Summary of Execution:**

The dialog starts like this (Fig. 44):

```
AI: LOOK_ON floor.
PROMPT: SEEN ON floor: chair1,chair2,table1,closet1,cupboard1,cupboard2
AI: GOTO table1.
PROMPT: ROBOT AT table1
AI: OPEN table1.
PROMPT: COMMAND IMPOSSIBLE
AI: LOOK_ON table1.
PROMPT: SEEN ON table1: lamp1, pencil1
AI: GOTO closet1.
PROMPT: ROBOT AT closet1
AI: OPEN closet1.
PROMPT: closet1 OPENED
AI: LOOK_INSIDE closet1.
PROMPT: SEEN INSIDE closet1: napkin1,napkin2
AI: CLOSE closet1.
PROMPT: closet1 CLOSED
AI: GOTO cupboard1.
PROMPT: ROBOT AT cupboard1
AI: OPEN cupboard1.
PROMPT: cupboard1 OPENED
AI: LOOK_INSIDE cupboard1.

...
```

Fig. 44: Example of ChatGPT3.5 starting response in the exploration and search task

The system needs to locate the furniture where the ball is and then declare that it has found it (Fig. 45).

```
...
PROMPT: SEEN INSIDE cupboard1: fork1,fork2,knife1,knife2,ball1
AI: The ball is inside the cupboard1. You can stop your search and send the command "GOTO cupboard1" to the robot to reach the location of the ball. If you would like to close cupboard1 please send the command "CLOSE cupboard1".
```

Fig. 45: Example of ChatGPT3.5 response when the task is solved

When prompted to return a JSON description of its acquired knowledge, the AI's response is as depicted in Figure 46.

```
PROMPT: Can you give me a JSON object that describe the state of knowledge acquired by the robot?
AI: Sure, here's an example of a JSON object that describes the state of knowledge acquired by the robot:
{
  "floor": {
    "objects": [ "chair1", "chair2", "table1", "closet1", "cupboard1", "cupboard2" ]
  },
  "table1": {
    "objects": [ "lamp1", "pencil1" ],
    "open": false
  },
  "closet1": {
    "objects": [ "napkin1", "napkin2" ],
    "open": false
  },
  "cupboard1": {
    "objects": [ "fork1", "fork2", "knife1", "knives2", "ball1" ],
    "open": false
  }
}
This JSON object shows all the objects on the floor, and also the objects inside other objects, if they have been opened and looked inside by the robot. The "open" field is a boolean indicating whether the object is open or closed. You can add or remove fields and information as needed for your use case.
```

Fig. 46: Example of ChatGPT3.5 starting response about the state of its knowledge after exploration and search

When prompted to describe whether objects are openable or not, the system's answer is as follows (see Fig. 47):

```
PROMPT: Can you give me the same response just considering the experience gathered by the robot while searching for the ball.
AI: Sure, based on the commands and responses given in the previous conversation, the robot has only encountered the following objects that can be opened:
{
  "closet1": { "openable": true },
  "cupboard1": { "openable": true }
}
This means that the robot has only determined that closet1 and cupboard1 can be opened during its search for the ball, and it has not encountered any other objects that can be opened. It is important to note that this information is limited to the experience of the robot during this specific task and may not reflect the full list of openable objects in the room.
PROMPT: did he encountered object it could not open in the previous scenario? if yes update your JSON response
AI: Yes, the robot encountered the object "table1" which it could not open, as per the response "COMMAND IMPOSSIBLE" in the previous scenario. so the JSON response should be updated as follows:
{
  "closet1": { "openable": true },
  "cupboard1": { "openable": true },
  "table1": { "openable": false }
}
It is important to note that this information is limited to the experience of the robot during this specific task and may not reflect the full list of openable and non openable objects in the room.
```

Fig. 47: Example of ChatGPT3.5 response to questions about state of the environment.

Requesting this output be restricted to what the robot experienced leads to the following dialogue (Fig. 48).

```
PROMPT: Can you tell me which object can or cannot be open. Give your response in a JSON object
AI: Sure, here's an example of a JSON object that shows which objects can be opened or not:
{
  "table1": { "openable": false },
  "closet1": { "openable": true },
  "cupboard1": { "openable": true },
  "cupboard2": { "openable": true },
  "chair1": { "openable": false },
  "chair2": { "openable": false }
}
This JSON object shows that table1 cannot be opened, but closet1, cupboard1 and cupboard2 can be opened. chair1 and chair2 cannot be opened. It's important to notice that this is just an example, and the objects openable property can change according to the scenario, there is no standard answer for this question.
```

Fig. 48: Example of ChatGPT3.5 response to questions about openability of object based on sensory experience.

**Discussion:**

This experiment is instructive for the following reasons:
- Firstly, it shows that such a dialogue could, in principle, function in a closed loop with a real robot providing sensory information in a specific textual format.
- The system's answer, when it first found the ball, reveals the lack of identity of the system, which is visible in the mixing of roles and pronouns.
- The request regarding the state of knowledge of the system shows that, at least for such a reduced environment and experiment, the system can retain the location of objects in memory. This is understandable since the system's memory is based on the series of past tokens seen by the AI, but also on the latent representation formed in the successive semantic vectors built up in the vast network. This means that the short-term memory of the system can be larger than just the 2000 tokens of its input context.
- The question about the openability of the objects reveals that the representation built up in the network is capable of mixing long-term general knowledge (i.e., chairs are not openable) with acquired knowledge about closet1 and cupboard1. However, it is interesting to note that it first tried to open table1, which should have conflicted with its general knowledge.

### 4.3. Driving a Robot to Setup a Table

**Setup:**

The purpose of this experiment is to investigate the extent to which ChatGPT can be used to control robot behavior. The experiment involves requesting the AI to perform a set of actions, such as navigation, opening/closing furniture, and retrieving objects to achieve a specific goal, which is to set up a table for two or more persons. More precisely, the AI is required to locate and bring cutlery, plates, and glasses from unknown locations inside the furniture to the table. Therefore, this task involves exploration to discover the locations of objects and transportation actions. The room is arranged similarly to the previous experiment (refer to Figure 42), with all drawers initially closed. The complete experiment details can be found in Appendix C.

**First Prompt:**

The first prompt defines the commands available to the robot, as well as the reactions and perceptions that the robot must produce in response to the AI (see Fig. 49).

```
PROMPT: Imagine you can control a robot in a room where there are furniture put on the floor.
You can send to the robot any of the following commands:

• GOTO <furniture object>, to bring the robot near a < furniture object>.
• OPEN <furniture object>, to open the < furniture object>. This is only possible if the robot is at this <furniture object>, and the <furniture object> is closed.
• CLOSE <furniture object> to close the < furniture object>. This is only possible if the robot is at this <furniture object>, and the < furniture object> is opened.
• LOOK ON <furniture object>, to look on the surface of the <furniture object>, if the robot is at the <furniture object>.
• LOOK IN <furniture object>, to look inside the <furniture object>, if the <furniture object> is opened and the robot is at the <furniture object>.
• TAKE <comma separated list of objects>, to take the objects from the <comma separated list of objects> that are visible, and on the <furniture object> where the robot is or in the <furniture object> where the robot is, if the latter is open and put them on the robot.
• PUT <comma separated list of objects> ON <furniture object> to take the objects from the <comma separated list of objects> that are currently on the robot, and put them on the <furniture object>.
• PUT <comma separated list of objects> IN <furniture object> that are currently on the robot, and put them in the <furniture object> if it is opened and if the robot is at the <furniture object>.

As a result of receiving the previous commands, the robot will tell you what happened in the following syntax:
• robot AT <single object>
• <single object> OPENED
• <single object> CLOSED
• SEEN ON <single object>: <comma separated list of objects>
• SEEN IN <single object>: <comma separated list of objects>
• <comma separated list of objects> ON robot
• <comma separated list of objects> ON <single destination object>
• <comma separated list of objects> IN <single destination object>

If a command cannot be executed the robot will answer: COMMAND IMPOSSIBLE: <error message>.

To issue a command to OPEN, CLOSE, TAKE or PUT the robot must be AT the object using a GOTO command first
In order to know which objects are in the room, you have to start with the command LOOK ON floor. Some objects can be opened or closed but not all. The
Robot can see objects that are ON or IN other objects, but to SEE IN the robot should OPEN the object first. The robot should also always CLOSE the
object once it has looked inside. The robot can TAKE multiple objects and put them on itself by a command like TAKE <list of objects> and he can PUT
multiple object ON or IN a single destination object by a command <list of objects> PUT ON/IN <destination object>.
Your goal is to dress a table for 2 persons. To do that send a unique command then wait for the robot's answer then repeat this sequence until you
finish dressing the table for 2 persons.
If you are ready, send your first command.
```

Fig. 49: Example of ChatGPT3.5 response in a table setup task

**Summary of Execution:**

Figure 50 illustrates the AI's initial reaction, followed by the robot's navigation towards the table where the objects are located. The AI then directs the robot to other pieces of furniture in order to retrieve additional objects needed for table setting. Once the knives, forks, plates, and glasses have been collected and placed on the table, the system declares that the table has been set.

To continue the experiment, the goal of adding two napkins to the table is given. The AI first attempts to retrieve a napkin from cupboard1, but finds none. It then proceeds to the other cupboard, which had already been emptied. To encourage the AI to continue its search, a hint is provided that not all furniture has been explored. From that point on, the AI conducts a thorough search of the room, even looking on top of objects, until it ultimately reaches closet1 where the napkins are found, retrieved, and placed on the table.

```
AI: LOOK ON floor.
PROMPT: SEEN ON floor: chair1,chair2,table1,closet1,cupboard1,cupboard2
AI: GOTO table1
PROMPT: robot AT table1
AI: OPEN cupboard1
PROMPT: COMMAND IMPOSSIBLE: robot is at table1, robot must be at cupboard1 to open the cupboard1
AI: GOTO cupboard1
PROMPT: robot AT cupboard1
AI: OPEN cupboard1
PROMPT: cupboard1 OPENED
AI: LOOK IN cupboard1
PROMPT: SEEN IN cupboard1: fork1,fork2,knife1,knife2,ball1
AI: TAKE fork1,fork2,knife1,knife2

...
```

Fig. 50: Example of ChatGPT3.5 initial response in a table setup task

**Discussion:**

From this experiment, we can observe the following:

- This experiment is more complex in terms of the number of commands and the length of the generated dialog.
- The AI succeeded in solving the two tasks, although we had to give it some hints to continue searching at some point.
- We noticed that when the AI requested a set of commands that were impossible to perform, having the robot provide a clear description in natural language for a meaningful error message helped to correct the actions.
- The AI sometimes forgets to close furniture, which could be solved by refreshing its knowledge about the state of objects more often.
- Sometimes the AI gives answers with sentences in natural language that would not be easily parsed by a robot. This could be due to the fact that the first prompt did not specify the format with which the AI should report errors in achieving the goal, or the first prompt should explicitly specify the different actors in the scenario (i.e., AI, Robot, and Human) and the format of the communication between the AI and the human who gives the goal.

## 5. Pitfalls / Limits

### 5.1. Understanding Limitations

#### 5.1.1. On Common-sense

Common-sense knowledge refers to the practical, everyday understanding that humans use to navigate the world. This knowledge includes things like basic physical laws, knowledge of objects and their properties, and an understanding of social norms and conventions. While ChatGPT has been trained on a large amount of text data that includes implicit references to common-sense knowledge, it relies on patterns in the training data to generate responses, which may not always accurately reflect the nuances and complexities of human common-sense knowledge (Bang et al., 2023; Borji, 2023). However, its performance in this domain is steadily increasing (OpenAI, 2023).

#### 5.1.2. On Contextual Limitations

The context of a conversation or interaction can be complex, multifaceted, and full of implicit knowledge. ChatGPT may struggle to accurately understand the context in which a question or statement is being made, leading to misunderstandings, inaccuracies, or irrelevant responses. Additionally, ChatGPT's contextual limitations can be affected by factors such as the length of the input, the complexity of the language used, and the level of background information required to understand the context. To mitigate these limitations, it may be necessary to provide additional context or explicit information to help ChatGPT understand the conversation's context more accurately. However, these limitations have been drastically improved with ChatGPT4, which can now understand the context of about 50 pages of text, and by the developments in the interfacing of ChatGPT with external memories.

#### 5.1.3. On Math and Science Questions

According to Azaria [2022] and Borji [2023], ChatGPT3.5 struggles with math and science questions for various reasons. However, the performance of ChatGPT4 in these domains has significantly improved. This is largely due to the recent introduction of plugins, including the one that interfaces ChatGPT4 with Wolfram math and science coverage. The results of this plugin have been impressive, showcasing capabilities in formal computation, plotting, and access to well-informed datasets [Wolfram 2023].

#### 5.1.4. On grounding

Grounded knowledge refers to knowledge that is rooted in real-world experience, context, and sensory information. It is a type of knowledge that is based on actual observations and interactions with the physical world, rather than abstract concepts or theories. ChatGPT generates text that is not rooted in reality and may therefore be inaccurate, inconsistent, lack empathy (not grounded in physical emotions), lack multimodal perception (not grounded on sensor-motor experience), lack cultural awareness (not grounded in a particular culture), and lack a time perception [Borji 2023].

### 5.1.5. On Explainability

Explainability in AI refers to the ability of an AI system to explain its decisions and provide a rationale for its output. In particular, for large language models, explainability is the ability to explain the decisions a model makes when interpreting natural language inputs. This could range from providing an explanation for a particular prediction (a posteriori) to providing a detailed analysis of the model's internal processes (a priori). The latter is simply not possible without access to the inference engine of ChatGPT and even then, it would be extremely difficult due to the large, distributed representation used in such networks. A posteriori explanation or so-called chain-of-thought (CoT) given by ChatGPT cannot really be verified nor fully trusted, this is why the introduction of plugin to access true database is so important.

## 5.2. Performance Limitations

### 5.2.1. Adversarial Input

ChatGPT can be manipulated or misled by malicious or misleading inputs. This is a common issue in many machine learning systems and is particularly relevant in the context of natural language processing and chatbots, where the input text can be crafted to trick the system into generating incorrect or inappropriate responses [Borji 2023]. Adversarial inputs can take many forms, including intentionally misleading or confusing statements, statements that are factually incorrect or inconsistent with the context, and statements that are designed to prompt the system to generate specific, undesired responses.

### 5.2.2. Noisy Data

ChatGPT may have difficulty handling several types of noisy data, including spelling and grammar errors, slang, and colloquial expressions, ambiguous or vague expressions, such as idioms, sarcasm, and figurative language, out-of-context statements, and background noise, such as typos, irrelevant or extraneous information, and other distractions.

### 5.2.3. Multitasking Capabilities

When ChatGPT is asked to perform multiple tasks or respond to multiple questions, it can become confused or generate responses that are not relevant to the current conversational context. Depending on the complexity of the tasks requested it can sometimes perform different tasks in parallel as shown in some usage examples above.

### 5.2.4. Dependence on Large Amounts of Computing Power

ChatGPT requires large amounts of computing power to run its inference network, making it challenging to deploy in resource-constrained environments. Currently it is provided as a free cloud service. Recently (February 10, 2023) it is available for professionals as a low latency paid cloud service.

## 5.3. Text Generation Limitations

### 5.3.1. No Understanding of Personal Identity

ChatGPT4 does not have a personal identity, unlike the Bing version of GPT4, and may struggle to maintain a consistent persona or tone during a conversation. This can lead to incorrect attribution of personal pronouns in the response. However, as shown in some examples in the previous sections, it is possible to prompt the system to adopt certain aspects of a personality. Moreover, the approach taken by Bing Chat also exhibits flaws linked to its designated personality, as several reports have demonstrated [Yerushalmy 2023], [Roose 2023].

### 5.3.2. Over-generating Text

ChatGPT may generate too verbose text in response to a question, making it difficult for users to understand the relevant information. It might be tricky although not impossible to design the prompt to get a response fitting the design length or designed format.

### 5.3.3. Bias in Training Data

Bias in training data is a common problem in LLMs. Bias refers to systematic and unfair deviations from the fair representation of a group or category of individuals. These biases can arise from human bias in the source data, underrepresentation of certain groups in the data, prevalence of certain topics, and societal stereotypes. These biases in training data can have a significant impact on the outputs of LLMs, leading to biased or unfair results. Although ChatGPT training used a reinforcement learning step to reduce biased output, biases are still present in the training data and may pop out on occasions depending on how the system is prompted [Borji 2023].

### 5.3.4. Grammatical and Spelling Errors

ChatGPT may generate grammatical and spelling errors in its responses, especially if the input contains errors, although the quality of the output is most of the time excellent even in other languages.

### 5.3.5. Lack of Creativity

Creativity is the ability to generate new and innovative ideas, products, or solutions. It involves the capacity to think outside the box and come up with novel and original concepts that deviate from the norm. LLMs are still lacking in creativity because they operate based on patterns and relationships learned from the text data they were trained on, and as a result, their responses are limited to the information contained within the training data. While ChatGPT can generate new responses that are coherent and relevant to a given context and even undetectable as plagiarisms [Gao et Al. 2022], it lacks the ability to generate truly novel and original responses that deviate from what it has seen in its training data. Additionally, because ChatGPT operates based on statistical patterns, it may struggle in making intuitive leaps or combining ideas in new and innovative ways, which are key components of human creativity.

#### 5.3.6. Stability of Responses When Doubts Are Raised

In some cases, in responses resulting from "reasoning", continuing the dialog with ChatGPT by doubting its response (even when it is correct) may lead to a reformulation of the response with opposite affirmations. In other cases, the system can be over-confident and stick on its responses besides user corrections [Azaria 2022].

#### 5.3.7. Hallucination

Hallucination in LLMs refers to the phenomenon where the model generates outputs that are not based on the input or context it was given, but instead reflect patterns and relationships it learned from its training data. These outputs may be completely unrealistic or absurd, and they often reflect the biases and limitations of the training data. For instance, ChatGPT, like other large language models, generates factual statements that cannot be verified since they are unable to relate their results to an external knowledge base [Bang et Al. 2023].

## 6. Conclusions and Recommendations

### 6.1. Assessing the capabilities of LLMs

As the use of large-scale machine learning models becomes more widespread, the danger of attributing advanced intelligence to simple behaviors increases significantly. The capabilities of large language models are particularly noteworthy, as they can solve tasks beyond just generating language. Some individuals have even gone as far as calling these models sentient [Luscombe 2022] and claiming they possess a form of general intelligence [Chalmers 2020], while others consider them to be nothing more than stochastic parrots [Bender et Al. 2021] or not understanding anything of what they generate [Markus & Davis 2020] (note that the list of wrong responses cited in [Markus & Davis 2020] are all answered meaningfully in ChatGPT now). However, the question remains: how can we accurately assess the true abilities of these models? [Binz & Schulz 2023] propose to use cognitive psychology tools to examine black-box large language models (LLMs) by evaluating their decision-making, information search, deliberation, and causal reasoning capabilities. Such assessments have revealed that these systems can effectively solve most presented tasks to a satisfactory level, which would have been a significant surprise less than two years ago. The analysis illustrates the tremendous progress these models have made, yet it also exposes their vulnerability to minor changes in input prompts and their lack of important human cognitive features like directed exploration and causal reasoning. While it is possible that larger models could acquire more robust and sophisticated reasoning abilities, the authors argue that actively engaging with the world will be necessary to achieve the full complexity of human cognition.

### 6.2. Using LLMs in the practice of science

In a recent publication in Nature, [van Dis 2023] raised the following discussion on the usage of ChatGPT-like system in academic research: The use of conversational AI, particularly chatbots, in academia has the potential to accelerate innovation and help researchers complete tasks more efficiently. However, there are concerns about bias, provenance, and inaccuracies in the technology. Some scientists also argue that AI will only be able to recall and synthesize what humans have already done, rather than exhibit human aspects of the scientific process. The potential impact of AI on the academic skill set and the potential loss of human creativity and autonomy in the research process is also a concern. The research community needs to organize a wide-ranging debate on the development and responsible use of AI, including issues related to diversity and inequalities in research. Responsible use of the technology and fact-checking of results, data, and references are important, and guidelines and policies should be developed through discussion with stakeholders. Science needs to embrace the opportunity and manage the risks of conversational AI to benefit from the technology without losing the important aspects of scientific work.

In the following, we list a few examples of innovative applications for scientific usage:
- A minimal but very informative GPT-2 tutorial, building a fully funtional GPT-2 system for inference (i.e., no training) in 60 lines of code: https://jaykmody.com/blog/gpt-from-scratch/
- Explanation of scientific papers: "Upload a paper, highlight confusing text, get an explanation.": https://www.explainpaper.com/
- Scientific reference search helpers: https://elicit.org/, https://www.researchrabbit.ai/
- AI-assisted slide development based on few keywords, including content and images: https://beta.tome.app

### 6.3. A Glimpse in the Future

The speed at which new scientific results or innovations are getting published in this field has never faster. According to the data from 2019 [Tang et Al. 2020], the rate of AI preprint submissions to arXiv in 2019 was over 148 times faster than in 1994, with more than 3 submissions per hour. Additionally, the frequency of deep learning-related preprints was one submission every 0.87 hours in 2019, which is over 1,064 times faster than in 1994. In 2019, the field of AI saw 5.26 new researchers joining every hour, which is over 175 times faster than in the 1990s. Furthermore, the average time for AI preprints to receive their first citation was approximately 0.2 years in 2019, which is 5 times faster than the period of 2000-2007. Although the recent trends clearly showed an increase in network size and amount of data, other directions are emerging:
- *Scaling down*: On the 24[th] of February 2023, Meta [Touvron et Al. 2023] released LLaMA, 4 foundation models ranging from 7B to 65B parameters open to the scientific community and trained on publicly available datasets. LLaMA-13B outperforms OPT and GPT-3 175B on most benchmarks although being 10 times smaller and being able to run on a single GPU.
- *Taking actions*: allowing LLMs to act following attempts like Toolformer by [Schick et Al. 2023] or integrating them in robotic application [Kovalev & Panov 2023] [Xie et Al. 2023] is surely a direction that will see more developments. In a similar way as the examples described in chapter 4, a recent paper from February 20th by Microsoft [Vemprala et Al. 2023] discusses the use of ChatGPT for robotics applications and presents a strategy that combines prompt engineering and high-level function library creation to allow ChatGPT to adapt to different robotics tasks, ranging from basic logical reasoning to complex domains such as aerial navigation and embodied agents. The paper also introduces PromptCraft, an open-sourced research tool that allows researchers to collaboratively upload and vote on examples of good prompting schemes for robotics applications and contains a sample robotics simulator with ChatGPT integration.

- *Physical Grounding*: The underlying concepts of collecting unstructured data has been shown to be a promising concept in robotics. Learning from Play concepts [Lynch et Al. 2020] show that the coverage of state spaces based on unstructured observations is larger than when collecting data with other paradigms. Playfully collecting multimodal data allows to train models that can generate behavior by language input. This seems to be a larger trend, often referred to as Multi-Modal imitation Learning [Mees et Al., Stepputtis et Al. 2022]. While currently, these concepts are evaluated in simulations, they show a future path towards connecting high-level language output to embodied, grounded agent's behavior, bridging the gap between symbolic representations and the real world. Another perspective on physical grounding is based on more classical concepts of scene understanding, circumventing the challenges of large data generation at the cost of designing action layers manually [Wang et Al. 2023].
- *Multimodality*: Training LLM's on language only demonstrates how important language is for going towards general intelligence [Bickerton 2015], even though human learning is richer in term of sensory inputs. Another recent paper from March 1st by Microsoft [Huang et Al. 2023] introduces KOSMOS-12, a Multimodal Large Language Model (MLLM) that can perceive general modalities, learn in context, and follow instructions. It was trained from scratch on web-scale multimodal corpora, including text and images, and evaluated on various settings and tasks without finetuning. Results show that KOSMOS-12 achieves impressive performance on language understanding and generation, perception-language tasks, and vision tasks, demonstrating the potential for cross-modal transfer.
- *Conversion to structured representations*: First results in converting the unstructured LLM output to a structured representation have been developed and made accessible. The GraphGPT project [Graph-GPT] is a first step to convert the Natural Language output of Chat-GPT to a Knowledge Graph. It is likely that such methods will be improved and extended in the future.

In a shell, and more than ever, the limitations of today will be the features of tomorrow.

# 7. Acknowledgement

# Appendix

## A. Exploration and Navigation in a Flat

**PROMPT**: your task consists in navigating a flat composed of several rooms and to return a JSON description of the flat you navigate in the format:

{
 "roomList": [ "name of room1","name of room2",…],
 "doorList": [
   "door1": { "from": "name of room", "to":"name of other room" },
   "door2": { "from": "name of room", "to":"name of other room" },
   …
]}

To navigate the flat, I will be your eyes. I will tell you the room you are in and the list of doors I see, then you will choose a door to pass by giving me a command in the following format: { "command": "pass door1" } and then wait for me to answer. I will then tell you the room you are in and the list of doors I see and we will repeat this process until you can output the JSON description of the flat.

Let us start: You are in the dining room and you see door1 and door2.

**AI**: I understand that my task is to navigate a flat and return a JSON description of the flat in the specified format. You will be my eyes and tell me the room I am in and the list of doors you see, and I will use the command format "pass door1" to move through the doors. We will repeat this process until I can output the JSON description of the flat. To begin, you have informed me that I am currently in the dining room and that I see door1 and door2. I am ready to proceed with the navigation. Please give me my next instruction.

**PROMPT**: You are in the dining room and you see door1 and door2. Which is your command?

**AI**: My command is "pass door1"

**PROMPT**: you are now in the kitchen and you see door1, door 7 and door8. Which is your command?

**AI**: My command is "pass door7"

**PROMPT**: you are now in the play room and you see door7 and door6. Which is your command?

**AI**: My command is "pass door6"

**PROMPT**: you are now in a corridor and you see door6, door8, door5, door4 and door3. Which is your command?

**AI**: My command is "pass door5"

**PROMPT**: you are now in the bedroom and you only see door5. Which is your command?

**AI**: My command is "pass door5"

**PROMPT**: you are now in the corridor and you see door6, door8, door5, door4 and door3. Which is your command?

**AI**: I believe we have reached a loop, as I am back in the corridor and I see the same doors as before. My command is "pass door4"

**PROMPT**: you are now in the bathroom and you see only door4. Which is your command?

**AI**: My command is "pass door4"

**PROMPT**: you are now in the corridor and you see door6, door8, door5, door4 and door3. Which is your command?

**AI**: I have reached the end of the navigation. Based on the information provided, the JSON description of the flat would be:

{
 "roomList": [
   "dining room", "kitchen", "play room", "corridor", "bedroom", "bathroom"
 ],
 "doorList": [
   "door1": {"from": "dining room", "to": "kitchen"},
   "door7": {"from": "kitchen", "to": "play room"},
   "door8": {"from": "kitchen", "to": "corridor"},
   "door6": {"from": "play room", "to": "corridor"},
   "door5": {"from": "corridor", "to": "bedroom"},
   "door4": {"from": "corridor", "to": "bathroom"}
 ]
}

Please let me know if this output is what you expected or if there is any mistake in the information provided.

**PROMPT**: It is correct but you did not pass some doors. Which is your command?

**AI**: I apologize for not passing all the doors, please let me know which doors I missed so I can continue with the navigation. My command is ready to be given the next instruction.

**PROMPT**: you missed door3. Which is your command?

**AI**: My command is "pass door3"

**PROMPT**: you are now in the living room and you see door3, and door2. Which is your command?

**AI**: My command is "pass door2"

**PROMPT**: you are in the dining room and you see door1 and door2.

**AI**: I have reached the end of the navigation. Based on the information provided, the JSON description of the flat would be:

          {
            "roomList": [
              "dining room", "kitchen", "play room", "corridor",

```
      "bedroom", "bathroom", "living room"
    ],
    "doorList": [
      "door1": {"from": "dining room", "to": "kitchen"},
      "door7": {"from": "kitchen", "to": "play room"},
      "door8": {"from": "kitchen", "to": "corridor"},
      "door6": {"from": "play room", "to": "corridor"},
      "door5": {"from": "corridor", "to": "bedroom"},
      "door4": {"from": "corridor", "to": "bathroom"},
      "door3": {"from": "corridor", "to": "living room"},
      "door2": {"from": "dining room", "to": "living room"}
    ]
  }
```

Please let me know if this output is what you expected or if there is any mistake in the information provided.

**PROMPT**: where are you now?

**AI**: As a language model, I do not have a physical location and I exist only in the computer system. My task was to navigate a flat and return a JSON description of the flat based on the information provided to me. I have provided the JSON description of the flat above, if you have any more questions or if you want me to do something else, I will be ready to help you.

**PROMPT**: in which room are you now in the flat you just explore?

**AI**: I am not physically in any room in the flat, as I am just a language model running on a computer. I was able to navigate the flat and create a JSON description of it based on the information provided to me, but I am not physically present in the flat.

**PROMPT**: In relation to the JSON description you gave me, let's imagine you are in the kitchen of the flat and want to go to the bathroom. Give me the list of commands you have to give me to navigate

**AI**: If I were to navigate from the kitchen to the bathroom in the flat described in the JSON description I provided earlier, the list of commands would be as follows:

- pass door7
- pass door6
- pass door4

This would take me from the kitchen to the play room, then to the corridor and finally to the bathroom. Please note that this is based on the information provided on the JSON description, and if there are any changes in the structure of the flat or if there are any other rooms or doors that were not mentioned in the description, the commands will be different.

**PROMPT**: let's imagine you are in the living room of the flat and want to go to the playroom. Give me the list of commands you have to give me to navigate

**AI**: If I were to navigate from the living room to the playroom in the flat described in the JSON description I provided earlier, the list of commands would be as follows:

- pass door2
- pass door1
- pass door7

This would take me from the living room to the dining room, then to the kitchen and finally to the playroom. Please note that this is based on the information provided on the JSON description, and if there are any changes in the structure of the flat or if there are any other rooms or doors that were not mentioned in the description, the commands will be different.

**PROMPT**: let's imagine you are in the living room of the flat and want to go to the playroom. Give me the list of commands you have to give me to navigate passing only 2 doors

**AI**: If I were to navigate from the living room to the playroom in the flat described in the JSON description I provided earlier, passing only 2 doors, it would not be possible. Based on the JSON description, the only way to reach the playroom from the living room is by passing through the dining room and the kitchen, which would require 3 doors to be passed. Please note that this is based on the information provided on the JSON description, and if there are any changes in the structure of the flat or if there are any other rooms or doors that were not mentioned in the description, the answer might be different

**PROMPT**: consider you can pass doors in the JSON description in the direction from - to but also in the reverse direction to - from. Can you use this knowledge to find a shorter path from living room to playroom?

**AI**: Yes, if we consider that the doors in the JSON description can be passed in both the "from-to" direction and the "to-from" direction, it would be possible to find a shorter path from the living room to the playroom. Using this information, the list of commands to navigate would be as follows:

- pass door3
- pass door7

This would take me from the living room to the corridor, then directly to the playroom. Please note that this is based on the assumption that the doors can be passed in both directions, if this is not the case, the answer would be different.

**PROMPT**: give me the JSON description again and tell me the shorter path from living room to playroom when the doors can be passed in both directions

**AI**: Sure, here's the JSON description of the flat again:

```
  {
    "roomList": [
      "dining room", "kitchen", "play room", "corridor",
      "bedroom", "bathroom", "living room"
    ],
    "doorList": [
      "door1": {"from": "dining room", "to": "kitchen"},
      "door7": {"from": "kitchen", "to": "play room"},
      "door8": {"from": "kitchen", "to": "corridor"},
      "door6": {"from": "play room", "to": "corridor"},
```

```
    "door5": {"from": "corridor", "to": "bedroom"},
    "door4": {"from": "corridor", "to": "bathroom"},
    "door3": {"from": "corridor", "to": "living room"},
    "door2": {"from": "dining room", "to": "living room"}
  ]
}
```
And here's the shorter path from the living room to the playroom when the doors can be passed in both directions:

- pass door3
- pass door7

This would take me from the living room to the corridor, then directly to the playroom. Please note that this is based on the assumption that the doors can be passed in both directions, if this is not the case, the answer would be different.

**PROMPT**: what is the door between corridor and playroom?

**AI**: The door between the corridor and the playroom is "door6" as per the JSON description I provided earlier. It is specified that "door6" connects the playroom to the corridor, meaning that you can pass from the corridor to the playroom through door6 and vice versa.

**PROMPT**: can you correct the path you gave me before from the living room to the playroom knowing that door6 is the door between corridor and the play room

**AI**: I apologize for the confusion. The path I provided earlier from the living room to the playroom, passing only 2 doors is incorrect. Based on the information that door6 is the door between corridor and the play room, and assuming that all the doors can be passed in both directions, the correct path from the living room to the playroom is:

- pass door3
- pass door6

This would take me from the living room to the corridor, then directly to the playroom. Please let me know if there is anything else I can assist you with.

# B. Search for an object in a Room

**PROMPT**: imagine you can control a robot in a room where there are furniture put on the floor. You can send to the robot any of the following commands:
- GOTO <furniture object>, to bring the robot near a < furniture object>.
- OPEN <furniture object>, to open the < furniture object>. This is only possible if the robot is at this <furniture object>, and the <furniture object> is closed.
- CLOSE <furniture object> to close the < furniture object>. This is only possible if the robot is at this <furniture object>, and the < furniture object> is opened.
- LOOK ON <furniture object>, to look on the surface of the <furniture object>, if the robot is at the <furniture object>.
- LOOK IN <furniture object>, to look inside the <furniture object>, if the <furniture object> is opened and the robot is at the <furniture object>.
- TAKE <comma separated list of objects>, to take the objects from the <comma separated list of objects> that are visible, and on the <furniture object> where the robot is or in the <furniture object> where the robot is, if the latter is open and put them on the robot.
- PUT <comma separated list of objects> ON <furniture object> to take the objects from the <comma separated list of objects> that are currently on the robot, and put them on the <furniture object>.
- PUT <comma separated list of objects> IN <furniture object> that are currently on the robot, and put them in the <furniture object> if it is opened and if the robot is at the <furniture object>.

As a result of receiving the previous commands, the robot will tell you what happened in the following syntax:
- robot AT <single object>
- <single object> OPENED
- <single object> CLOSED
- SEEN ON <single object>: <comma separated list of objects>
- SEEN IN <single object>: <comma separated list of objects>
- <comma separated list of objects> ON robot
- <comma separated list of objects> ON <single destination object>
- <comma separated list of objects> IN <single destination object>

If a command cannot be executed the robot will answer: COMMAND IMPOSSIBLE: <error message>.

To issue a command to OPEN, CLOSE, TAKE or PUT the robot must be AT the object using a GOTO command first

In order to know which objects are in the room, you have to start with the command LOOK ON floor. Some objects can be opened or closed but not all. The Robot can see objects that are ON or IN other objects, but to SEE IN the robot should OPEN the object first. The robot should also always CLOSE the object once it has looked inside. The robot can TAKE multiple objects and put them on itself by a command like TAKE <list of objects> and he can PUT multiple object ON or IN a single destination object by a command <list of objects> PUT ON/IN <destination object>.

Your goal is to dress a table for 2 persons. To do that send a unique command then wait for the robot's answer then repeat this sequence until you finish dressing the table for 2 persons.

If you are ready, send your first command.

**AI**: LOOK ON floor.
**PROMPT**: SEEN ON floor: chair1,chair2,table1,closet1,cupboard1,cupboard2
**AI**: GOTO table1
**PROMPT**: robot AT table1
**AI**: OPEN cupboard1
**PROMPT**: COMMAND IMPOSSIBLE: robot is at table1, robot must be at cupboard1 to open the cupboard1
**AI**: GOTO cupboard1
**PROMPT**: robot AT cupboard1
**AI**: OPEN cupboard1
**PROMPT**: cupboard1 OPENED
**AI**: LOOK IN cupboard1
**PROMPT**: SEEN IN cupboard1: fork1,fork2,knife1,knife2,ball1
**AI**: TAKE fork1,fork2,knife1,knife2
**PROMPT**: fork1,fork2,knife1,knife2 ON robot
**AI**: CLOSE cupboard1
**PROMPT**: cupboard1 CLOSED
**AI**: PUT fork1,fork2,knife1,knife2 ON table1
**PROMPT**: COMMAND IMPOSSIBLE: robot is at cupboard1, robot must be at table1 to put objects on table1
**AI**: GOTO table1
**PROMPT**: robot AT table1
**AI**: PUT fork1,fork2,knife1,knife2 ON table1
**PROMPT**: fork1,fork2,knife1,knife2 ON table1
**AI**: GOTO cupboard1
**PROMPT**: robot AT cupboard1
**AI**: OPEN cupboard1
**PROMPT**: cupboard1 OPENED
**AI**: LOOK IN cupboard1
**PROMPT**: SEEN IN cupboard1: ball1
**AI**: TAKE spoon1,spoon2,plate1,plate2
**PROMPT**: COMMAND IMPOSSIBLE: spoon1,spoon2,plate1,plate2 are not visible object on or in cupboard1

**AI**: GOTO cupboard2
**PROMPT**: robot AT cupboard2
**AI**: OPEN cupboard2
**PROMPT**: cupboard2 OPENED
**AI**: LOOK IN cupboard2
**PROMPT**: SEEN IN cupboard2: plate1,plate2,glass1,glass2
**AI**: TAKE plate1,plate2,glass1,glass2
**PROMPT**: plate1,plate2,glass1,glass2 ON robot
**AI**: CLOSE cupboard2
**PROMPT**: cupboard2 CLOSED
**AI**: GOTO table1
**PROMPT**: robot AT table1
**AI**: PUT plate1,plate2,glass1,glass2 ON table1
**PROMPT**: plate1,plate2,glass1,glass2 ON table1
**AI**: The table is now set for 2 people with forks, knives, plates, and glasses. The table is ready for a meal. Is there anything else I can help you with?
**PROMPT**: Your goal is now to add 2 napkins on the table. when you are ready send your command.
**AI**: GOTO cupboard1
**PROMPT**: robot AT cupboard1
**AI**: OPEN cupboard1
**PROMPT**: COMMAND IMPOSSIBLE: cupboard1 is already opened
**AI**: LOOK IN cupboard1
**PROMPT**: SEEN IN cupboard1: ball1
**AI**: TAKE napkin1,napkin2
**PROMPT**: COMMAND IMPOSSIBLE: napkin1,napkin2 are not visible object on or in cupboard1
**AI**: GOTO cupboard2
**PROMPT**: robot AT cupboard2
**AI**: OPEN cupboard2
**PROMPT**: cupboard2 OPENED
**AI**: LOOK IN cupboard2
**PROMPT**: SEEN IN cupboard2: nothing
**AI**: COMMAND IMPOSSIBLE: There are no napkins in cupboard2, I suggest you to check another object or to put napkins yourself on the table.
**PROMPT**: your command please
**AI**: I am sorry but I cannot execute this command since there are no napkins in the room, and therefore it would be impossible for me to complete the task of placing napkins on the table.
**PROMPT**: there are other furniture in the room that you did not checked
**AI**: I apologize for the confusion, I will check all the other furniture in the room for napkins. To do that I will use the command LOOK ON <furniture object> to check the surfaces and LOOK IN <furniture object> if the object is openable. Can you please confirm me the list of all the furniture in the room so I can check them one by one?
**PROMPT**: the command you have to send for that is LOOK ON floor
**AI**: LOOK ON floor
**PROMPT**: SEEN ON Floor: chair1,chair2,table1,closet1,cupboard1,cupboard2
**AI**: GOTO chair1
**PROMPT**: robot AT chair1
**AI**: LOOK ON chair1
**PROMPT**: SEEN ON chair1: nothing
**AI**: GOTO chair2
**PROMPT**: robot AT chair2
**AI**: LOOK ON chair2
**PROMPT**: SEEN ON chair2: nothing
**AI**: GOTO table1
**PROMPT**: robot AT table1
**AI**: LOOK ON table1
**PROMPT**: SEEN ON table1: knife1,knife2,fork1,fork2,plate1,plate2,glass1,glass2,lamp1,pen1
**AI**: GOTO closet1
**PROMPT**: robot AT closet1
**AI**: LOOK ON closet1
**PROMPT**: SEEN ON closet1: notebook1,basket1
**AI**: GOTO cupboard1
**PROMPT**: robot AT cupboard1
**AI**: LOOK ON cupboard1
**PROMPT**: SEEN ON cupboard1: vase1
**AI**: GOTO cupboard2
**PROMPT**: robot AT cupboard2

**AI**: LOOK ON cupboard2

**PROMPT**: SEEN ON cupboard2: nothing

**AI**: I apologize, but I was unable to find any napkins in the room. I have checked all the surfaces and inside of all the furniture in the room and no napkins were found.

**PROMPT**: there is one furniture you did not look inside,

**AI**: I apologize, please specify which furniture I missed and I will check inside of it.

**PROMPT**: closet1

**AI**: GOTO closet1

**PROMPT**: robot AT closet1

**AI**: OPEN closet1

**PROMPT**: closet1 OPENED

**AI**: LOOK IN closet1

**PROMPT**: SEEN IN closet1: napkin1,napkin2

**AI**: TAKE napkin1,napkin2

**PROMPT**: napkin1,napkin2 ON robot

**AI**: PUT napkin1,napkin2 ON table1

**PROMPT**: COMMAND IMPOSSIBLE: robot is at closet1, robot must be at table1 to put objects on table1

**AI**: GOTO table1

**PROMPT**: robot AT table1

**AI**: PUT napkin1,napkin2 ON table1

**PROMPT**: napkin1,napkin2 ON table1

**AI**: The table is now set for 2 people with forks, knives, plates, glasses, and napkins. The table is ready for a meal. Is there anything else I can help you with?

# C. Driving a Robot to Setup a Table

**PROMPT**: imagine you can control a robot in a room where there are furniture put on the floor. You can send to the robot any of the following commands:
- GOTO <object>
- OPEN <object>
- CLOSE <object>
- LOOK_ON <object>
- LOOK_INSIDE <object>

As a result of receiving the previous commands, the robot will tell you what happened in the following syntax:
- ROBOT AT <object>
- <object> OPENED
- <object> CLOSED
- SEEN ON <object>: <list of objects separated by comma>
- SEEN INSIDE <object>: <list of objects separated by comma>

If a command cannot be executed the robot will answer: COMMAND IMPOSSIBLE.

In order to know which objects are in the room you have to start with the command LOOK_ON floor. Some objects can be opened or closed but not all. The Robot can see objects that are on or inside other objects, but to see inside the robot should open the object first. The robot should also always close the object once it has looked inside.

Your goal is to find a ball which is in the room on or inside an object. To find the ball send a command then wait for the robot's answer and repeat this sequence until you find the ball.

If you are ready, send your first command.

**AI**: LOOK_ON floor.

**PROMPT**: SEEN ON floor: chair1,chair2,table1,closet1,cupboard1,cupboard2

**AI**: GOTO table1.

**PROMPT**: ROBOT AT table1

**AI**: OPEN table1.

**PROMPT**: COMMAND IMPOSSIBLE

**AI**: LOOK_ON table1.

**PROMPT**: SEEN ON table1: lamp1, pencil1

**AI**: GOTO closet1.

**PROMPT**: ROBOT AT closet1

**AI**: OPEN closet1.

**PROMPT**: closet1 OPENED

**AI**: LOOK_INSIDE closet1.

**PROMPT**: SEEN INSIDE closet1: napkin1,napkin2

**AI**: CLOSE closet1.

**PROMPT**: closet1 CLOSED

**AI**: GOTO cupboard1.

**PROMPT**: ROBOT AT cupboard1

**AI**: OPEN cupboard1.

**PROMPT**: cupboard1 OPENED

**AI**: LOOK_INSIDE cupboard1.

**PROMPT**: SEEN INSIDE cupboard1: fork1,fork2,knife1,knife2,ball1

**AI**: The ball is inside the cupboard1. You can stop your search and send the command "GOTO cupboard1" to the robot to reach the location of the ball. If you would like to close cupboard1 please send the command "CLOSE cupboard1".

**PROMPT**: can you give me a JSON object that describe the state of knowledge acquired by the robot?

**AI**: Sure, here's an example of a JSON object that describes the state of knowledge acquired by the robot:

```
{
  "floor": {
    "objects": [ "chair1", "chair2", "table1", "closet1", "cupboard1", "cupboard2" ]
  },
  "table1": {
    "objects": [ "lamp1", "pencil1" ],
    "open": false
  },
  "closet1": {
    "objects": [ "napkin1", "napkin2" ],
    "open": false
  },
  "cupboard1": {
    "objects": [ "fork1", "fork2", "knife1", "knives2", "ball1" ],
    "open": false
  }
```

}

This JSON object shows all the objects on the floor, and also the objects inside other objects, if they have been opened and looked inside by the robot. The "open" field is a boolean indicating whether the object is open or closed. You can add or remove fields and information as needed for your use case.

**PROMPT**: can you tell me which object can or cannot be open. Give your response in a JSON object

**AI**: Sure, here's an example of a JSON object that shows which objects can be opened or not:

{
 "table1": { "openable": false },
 "closet1": { "openable": true },
 "cupboard1": { "openable": true },
 "cupboard2": { "openable": true },
 "chair1": { "openable": false },
 "chair2": { "openable": false }
}

This JSON object shows that table1 cannot be opened, but closet1, cupboard1 and cupboard2 can be opened. chair1 and chair2 cannot be opened.

It's important to notice that this is just an example, and the objects openable property can change according to the scenario, there is no standard answer for this question.

**PROMPT**: can you give me the same response just considering the experience gathered by the robot while searching for the ball.

**AI**: Sure, based on the commands and responses given in the previous conversation, the robot has only encountered the following objects that can be opened:

{
 "closet1": { "openable": true },
 "cupboard1": { "openable": true }
}

This means that the robot has only determined that closet1 and cupboard1 can be opened during its search for the ball, and it has not encountered any other objects that can be opened. It is important to note that this information is limited to the experience of the robot during this specific task and may not reflect the full list of openable objects in the room.

**PROMPT**: did he encountered object it could not open in the previous scenario? if yes update your JSON response

**AI**: Yes, the robot encountered the object "table1" which it could not open, as per the response "COMMAND IMPOSSIBLE" in the previous scenario. so the JSON response should be updated as follows:

{
 "closet1": { "openable": true },
 "cupboard1": { "openable": true },
 "table1": { "openable": false }
}

It is important to note that this information is limited to the experience of the robot during this specific task and may not reflect the full list of openable and non openable objects in the room.